\documentclass[9pt,conference,letterpaper]{IEEEtran}
\usepackage{booktabs} 
\usepackage{amsmath,url,graphicx,amssymb}
\usepackage{amsfonts}
\usepackage{ctable}
\usepackage{multirow}
\usepackage{algorithm}
\usepackage{algpseudocode}
\usepackage{pifont}
\usepackage{color}
\usepackage{subfigure}
\usepackage{enumitem}
\usepackage{sidecap}
\usepackage{amsthm}

\newcommand{\mylistbegin}{
  \begin{list}{$\bullet$}
   {
     \setlength{\itemsep}{-2pt}
     \setlength{\leftmargin}{1em}
     \setlength{\labelwidth}{1em}
     \setlength{\labelsep}{0.5em} } }
\newcommand{\mylistend}{
   \end{list}  }

\newcommand{\eg}{\textit{e.g.}}

\newcommand{\ie}{\textit{i.e.}}
\newcommand{\etc}{\textit{etc}}

\newcommand{\wrt}{\textit{w.r.t.~}}
\newcommand{\header}[1]{{\vspace{+1mm}\flushleft \textbf{#1}}}

\newtheorem{theorem}{Theorem}[section]
\newtheorem{definition}[theorem]{Definition}

\title{Neural Embedding Propagation on Heterogeneous Networks}
\author{Anonymous Author(s)}
\author{
{Carl Yang$^*$, Jieyu Zhang$^*$, Jiawei Han}\\
\fontsize{10}{10}\selectfont\itshape
University of Illinois, Urbana Champaign, 201 N Goodwin Ave, Urbana, Illinois 61801, USA\\
\fontsize{9}{9}\selectfont\ttfamily\upshape
\{jiyang3, jieyuz2, hanj\}@illinois.edu
}

\begin{document}

\setlength{\floatsep}{4pt plus 4pt minus 1pt}
\setlength{\textfloatsep}{4pt plus 2pt minus 2pt}
\setlength{\intextsep}{4pt plus 2pt minus 2pt}
\setlength{\dbltextfloatsep}{3pt plus 2pt minus 1pt}
\setlength{\dblfloatsep}{3pt plus 2pt minus 1pt} 
\setlength{\abovecaptionskip}{3pt}
\setlength{\belowcaptionskip}{2pt}
\setlength{\abovedisplayskip}{2pt plus 1pt minus 1pt}
\setlength{\belowdisplayskip}{2pt plus 1pt minus 1pt}

\maketitle
{
\renewcommand{\thefootnote}{\fnsymbol{footnote}}
\footnotetext[1]{Both authors contributed equally to this work.}
}

\begin{abstract}
Classification is one of the most important problems in machine learning. To address label scarcity, semi-supervised learning (SSL) has been intensively studied over the past two decades, which mainly leverages data affinity modeled by networks. Label propagation (LP), however, as the most popular SSL technique, mostly only works on homogeneous networks with single-typed simple interactions. 
In this work, we focus on the more general and powerful heterogeneous networks, which accommodate multi-typed objects and links, and thus endure multi-typed complex interactions. 
Specifically, we propose \textit{neural embedding propagation} (NEP), which leverages distributed embeddings to represent objects and dynamically composed modular networks to model their complex interactions. While generalizing LP as a simple instance, NEP is far more powerful in its natural awareness of different types of objects and links, and the ability to automatically capture their important interaction patterns.
Further, we develop a series of efficient training strategies for NEP, leading to its easy deployment on real-world heterogeneous networks with millions of objects.
With extensive experiments on three datasets, we comprehensively demonstrate the effectiveness, efficiency, and robustness of NEP compared with state-of-the-art network embedding and SSL algorithms.
\end{abstract}
\section{Introduction}
\label{sec:intro}
The last decade has witnessed tremendous success of deep learning models, most of which highly rely on the availability of large amounts of training data \cite{he2016deep, simonyan2015very}. \textit{Semi-supervised learning} (SSL) \cite{zhu2002learning, zhu2003semi}, which is essentially close to the recent popular scheme of \textit{few-shot learning} \cite{vinyals2016matching, finn2017model, ravi2016optimization}, naturally aims at alleviating such reliance on training data.
Arguably the most classic model for SSL is based on transductive inference on graphs, \ie, \textit{label propagation} (LP) \cite{zhu2002learning, zhu2003semi}. Given a mixed set of labeled and unlabeled data points (\eg, images), LP firstly constructs a homogeneous affinity network (\eg, a $k$-nearest-neighbor adjacency matrix), and then propagates labels on the network. Due to its simplicity and effectiveness, LP has found numerous industrial applications \cite{baluja2008video, weston2012deep} and attracted various following-up research \cite{lin2014geodesic, yang2016revisiting, kipf2016semi}. 

While SSL is well studied on homogeneous networks, in the real world, however, data are often multi-typed and multi-relational, which can be better modeled by heterogeneous networks \cite{chen2017task, jiang2017semi}. For example, in a movie recommendation dataset \cite{yang2018meta}, the basic units can be \textsf{users}, \textsf{movies}, \textsf{actors}, \textsf{genres} and so on, whereas in a place recommendation dataset \cite{yang2018similarity, yang2017bridging} 
objects can be \textsf{users}, \textsf{places}, \textsf{categories}, \textsf{locations}, \etc. Moreover, knowledge bases such as \textsf{Freebase} and \textsf{YAGO} can also be naturally modeled by heterogeneous networks, due to their inherent rich types of objects and links. 
As a powerful model, heterogeneous network enables various tasks like the classification and recommendation of movies and places, as well as relational inference in knowledge bases, where SSL is highly desired due to the lack of labeled data.
However, trivial adoptions of LP on heterogeneous networks by suppressing the type information are not ideal, since they do not differentiate the functionalities of multiple types of objects and links.

To leverage the multi-relational nature of heterogeneous networks, the concept of \textit{meta-paths} (or \textit{meta-graphs}, as a more general notion) has been proposed and widely used by existing models on heterogeneous networks \cite{sun2011pathsim, shi2017prep, dong2017metapath2vec}. 
For the particular problem of SSL, \cite{jiang2017semi, ji2010graph, luo2014hetpathmine, chen2017task} have also leveraged meta-paths to capture the different semantics among targeted types of objects. 
However, as pointed out in \cite{wang2015knowsim, yang2018similarity}, 
the assumption that all useful meta-paths can be pre-defined by humans is often not valid, and exhaustive enumeration and selection over the exponential number of all possible ones is impractical. Therefore, existing methods considering a fixed set of meta-paths cannot effectively capture and differentiate various object interactions on heterogeneous networks. 

\begin{figure}[t]
 \centering\includegraphics[width=1\linewidth]{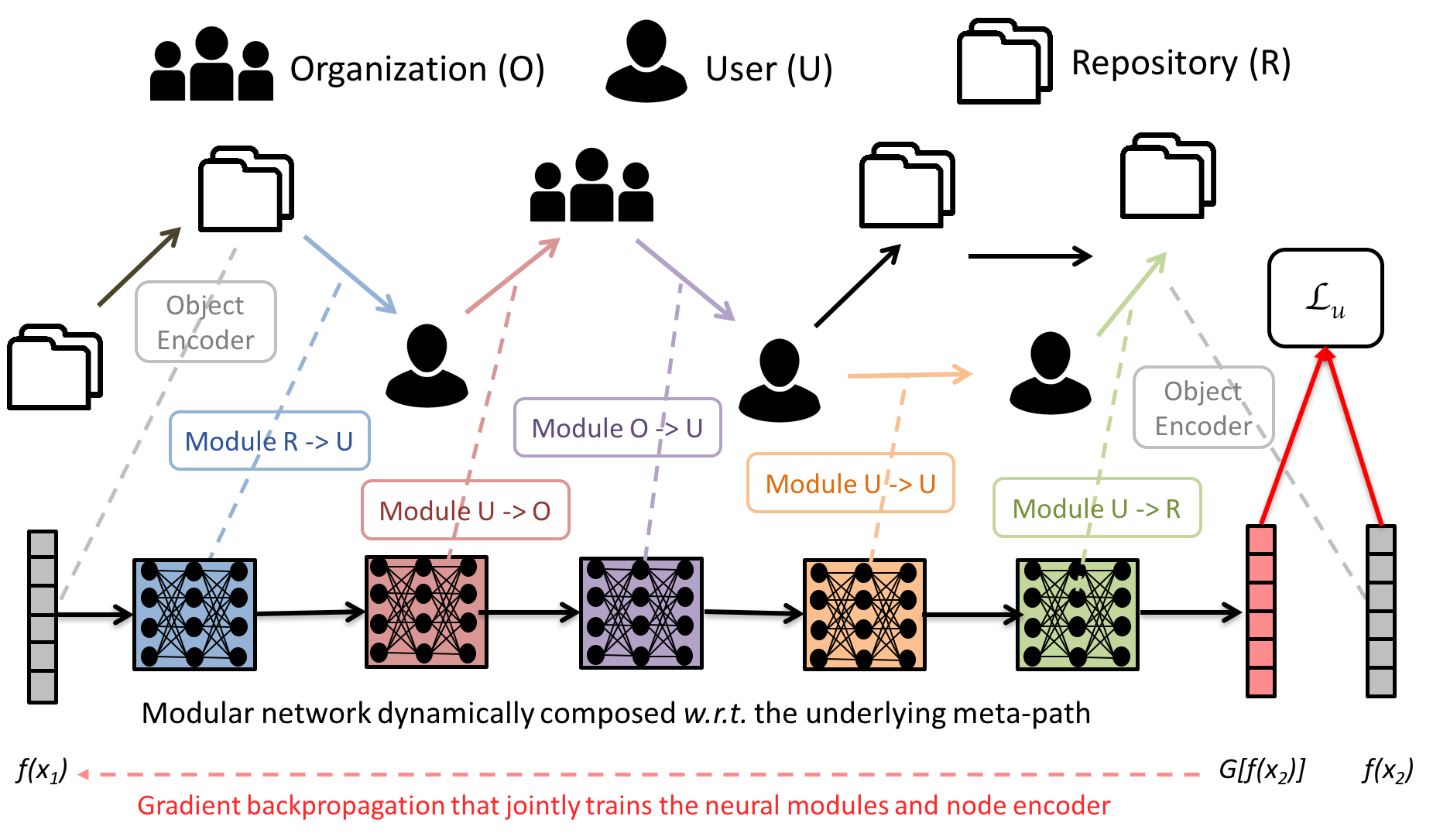}
 \caption{A running toy example of NEP on the GitHub heterogeneous network.}
 \label{fig:toy}
\end{figure}

In this work, to address the limitations of the existing works, we propose a novel NEP (\textit{Neural Embedding Propagation}) framework for SSL over heterogeneous networks. Figure \ref{fig:toy} gives a running toy example of NEP, which is a \textit{powerful} yet \textit{efficient} neural framework that coherently combines an \textit{object encoder} \cite{yang2018similarity, yang2017bridging}
and a \textit{modular network} \cite{andreas2016neural, andreas2016learning}. 
It leverages the compositional nature of meta-paths and trivially generalizes to attributed networks.

In Figure \ref{fig:toy}, the object encoder can be simply implemented as an embedding look-up table. Trained together with a parametric predictive model (\eg, a multi-layer perceptron (MLP)), it computes a mapping between object representations in a latent embedding space and object labels given as supervision. Such embeddings capture the correlations among object labels, and alleviates their inherent sparsity and noise. We find it also important to allow the embedding of labeled objects to change along training, which indicates that unlabeled data may even help improve the modeling of labeled data. 

One step further, to capture the complex interactions on different types of links, we cast each of them as a unique differentiable neural network module (\eg, also an MLP). Different meta-paths then correspond to unique modular networks, which are dynamically composed through stacking the corresponding neural network layers \wrt~the particular link types along the paths. During the training of NEP, each time starting from a particular object, to mimic the process of LP, we \textit{propagate} its \textit{label} along a particular sampled path, by feeding its \textit{object embedding} into the corresponding \textit{modular network}. An $\ell_{2}$-loss is computed between the \textit{propagated} embedding and the original embedding on the end object, to require proper \textit{smoothness} between the connected objects. Then the gradients are back propagated along the path to update both the corresponding neural modules and the object encoder.

Due to the expressiveness of neural networks, NEP is able to automatically discover the functionalities of different types of links and dynamically model their common compositions (\ie, meta-paths) on-the-fly based on uniform random walks, which allows us to abandon the explicit consideration of a limited set of meta-paths but rather model them in a data-driven way. Finally, as non-linearity can be easily added into the MLP-based neural network modules, NEP can be more flexible with complex object interactions.

To further improve the efficiency of NEP, we design a series of intuitive and effective training strategies. 
Firstly, in most scenarios, we only care about the labels of certain targeted types of objects. 
This allows us to only compute their embeddings and sample the random paths only among them. 
Secondly, to fully leverage training labels, we reversely sample the random paths from labeled objects, which makes sure the propagation paths all end on labeled objects, so the propagated embeddings can directly encode high-quality label information. 
Finally, to boost training efficiency, we design a two-step path sampling approach, which essentially groups instances of the same meta-paths into mini-batches, so that the same modular network is instantiated and trained in each mini-batch, leading to 300+ times gain on efficiency as well as slight gain on effectiveness.

Our experiments are done on three real-world heterogeneous networks with millions of objects and links, where we comprehensively study the effectiveness, efficiency and robustness of NEP. 
NEP is able to achieve $23.2\%-33.6\%$ relative gain on classification accuracy compared with the average scores of all baselines across three datasets, which indicates the importance of the proper modeling of complex object interactions on heterogeneous networks. Besides, NEP is also shown to be the most efficient regarding the leverage of training data and computational resources, while being robust towards hyper-parameters in large ranges.
All code will be released upon the acceptance of this work.

\section{Related Work and Preliminaries}
\label{sec:related}
\subsection{Heterogeneous Network Modeling}
Networks are widely adopted as a natural and generic model for interactive objects. 
Most of recent network models focus on the higher-order object interactions, since few interactions are independent of others. Arguably, the most popular ones include personalized page rank \cite{jeh2003scaling} and DeepWalk \cite{perozzi2014deepwalk} based on random walks, LINE \cite{tang2015line} and graph convolutional networks \cite{kipf2016semi} leveraging the direct node neighborhoods, as well as higher-order graph cut \cite{benson2016higher} and graph kernels methods \cite{shervashidze2011weisfeiler} considering small network motifs with exact shapes. All of them have stimulated various follow-up works, the discussion of which is beyond the scope of this work. 

In the real world, objects have multiple types and interact in different ways, which leads to the invention of heterogeneous networks \cite{sun2012mining}.
Due to its capacity of retaining rich representations of objects and links, it has drawn increasing research attention in the past decade and facilitated various downstream applications including link prediction \cite{liu2017semantic}, classification \cite{hou2017hindroid}, clustering \cite{sun2013pathselclus}, recommender systems \cite{zhao2017meta}, outlier detection \cite{zhuang2014mining} and so on. 

Since objects and links in heterogeneous networks have multiple types, the interaction patterns are much more complex. To capture such complex interactions, the tool of meta-path has been proposed and leveraged by most existing models on heterogeneous networks \cite{sun2011pathsim}. Traditional object proximity models measure the total strength of various interactions by counting the number of instances of different meta-paths between objects and adding up the counts with pre-defined or learned weights \cite{sun2011pathsim, wang2015knowsim, shi2017prep, wan2015graph, luo2014hetpathmine, ji2010graph, li2016transductive, jiang2017semi, fang2016semantic}, 
whereas the more recent network representation learning methods leverage meta-path guided random walks to jointly model multiple interactions in a latent embedding space \cite{dong2017metapath2vec, shang2016meta, yang2018meta, fu2017hin2vec, chen2017task, eswaran2017zoobp}. 
However, the consideration of a fixed set of meta-paths, while helping regulate the complex interactions, largely relies on the quality of the meta-paths under consideration, and limits the flexibility of the model, which is unable to handle any interactions not directly captured by the meta-paths. 

\subsection{Semi-Supervised Learning}
Semi-supervised learning (SSL) aims at leveraging both labeled and unlabeled data to boost the performance of various machine learning tasks. 
Among many SSL methods, the most classic and influential one might be label propagation (LP) \cite{zhu2002learning, zhu2003semi}. Its original version assumes the input of a small amount of labeled data and a data affinity network, either computed based on the distances among attributed objects, or derived from external data.
To predict the labels of unlabeled data, it propagates the labels from labeled data based on the topology of the affinity network, with the smoothness assumption that nearby objects on the network tend to have similar labels. Due to the simplicity and effectiveness of LP, many follow-up works have been proposed to improve it, especially on the homogeneous network setting \cite{lin2014geodesic, yang2016revisiting, kipf2016semi, yang2017bi}. 

SSL has also been studied in the heterogeneous network setting. The uniqueness of heterogeneous network is its accommodation of multi-typed objects and relations, thus leading to complex object interactions and propagation functions.
Therefore, all SSL models on heterogeneous networks leverage a given set of meta-paths to regulate and capture the complex object interactions. For example, \cite{ji2010graph, luo2014hetpathmine} both use a set of meta-paths to derive multiple homogeneous networks and optimize the label propagation process on all of them, whereas \cite{wan2015graph, jiang2017semi, serafino2018ensemble, yang2018did, shi2019user} jointly optimize the weight of different meta-paths. \cite{chen2017task, eswaran2017zoobp} simultaneously preserves the object proximities \wrt~multiple meta-paths to learn a unique network embedding. However, besides the limitation of given set of meta-paths, they still only consider simple interaction patterns with linear propagation functions.

\section{Neural Embedding Propagation}
\label{sec:model}
In this section, we describe our NEP (\textit{Neural Embedding Propagation}) algorithm, which coherently combines embedding learning and modular networks into a powerful yet efficient SSL framework over heterogeneous networks.

\subsection{Motivations and Overview}
In this work, we study SSL over heterogeneous networks. Therefore, the input of NEP is a heterogeneous network $\mathcal{G}=\{\mathcal{V,E}\}$, where $\mathcal{V}$ and $\mathcal{E}$ are the multi-typed objects and links, respectively. In general, $\mathcal{V}$ can be associated with $\{\mathcal{Y}, \mathcal{A}\}$, where object labels $\mathcal{Y}$ is often only available in a small subset $\mathcal{V}_l \subset\mathcal{V}$, and object attributes $\mathcal{A}$ can be available for all objects, part of all objects, or none of the objects at all. In this work, we focus on predicting the labels of all objects in $\mathcal{V}$ based on both $\mathcal{Y}$ and $\mathcal{E}$.

Before formally introducing the heterogeneous network setting, let us first consider SSL over homogeneous networks. Particularly, we aim to explain why LP is sufficiently effective in that situation. 

In homogeneous networks, since all objects share a single type, it is legitimate for LP to directly put any labels to any objects in the network. Also, labels on a single type of objects are often mutually exclusive and thus can be considered disjointly without a predictive model. Moreover, since all links share a single type, the only thing that can differ across links is their weight, which can be easily modeled by simple linear propagation functions. 

To understand the unique challenges of SSL in the heterogeneous network setting, we firstly briefly review the definition of heterogenous networks as follows.

\begin{definition}
A \textit{heterogeneous network} \cite{sun2012mining, sun2011pathsim} is a network $\mathcal{G}=\{\mathcal{V,E}\}$ with multiple types of objects and links. Within $\mathcal{G}$, $\mathcal{V}$ is the set of objects, where each object $v\in\mathcal{V}$ is associated with an object type $\phi(v)$, and $\mathcal{E}$ is the set of links, where each link $e\in\mathcal{E}$ is associated with a link type $\psi(e)$. It is worth noting that a link type automatically defines the object types on its two ends.
\label{def:hin}
\end{definition}

\begin{figure}[h]
 \centering\includegraphics[width=0.8\linewidth]{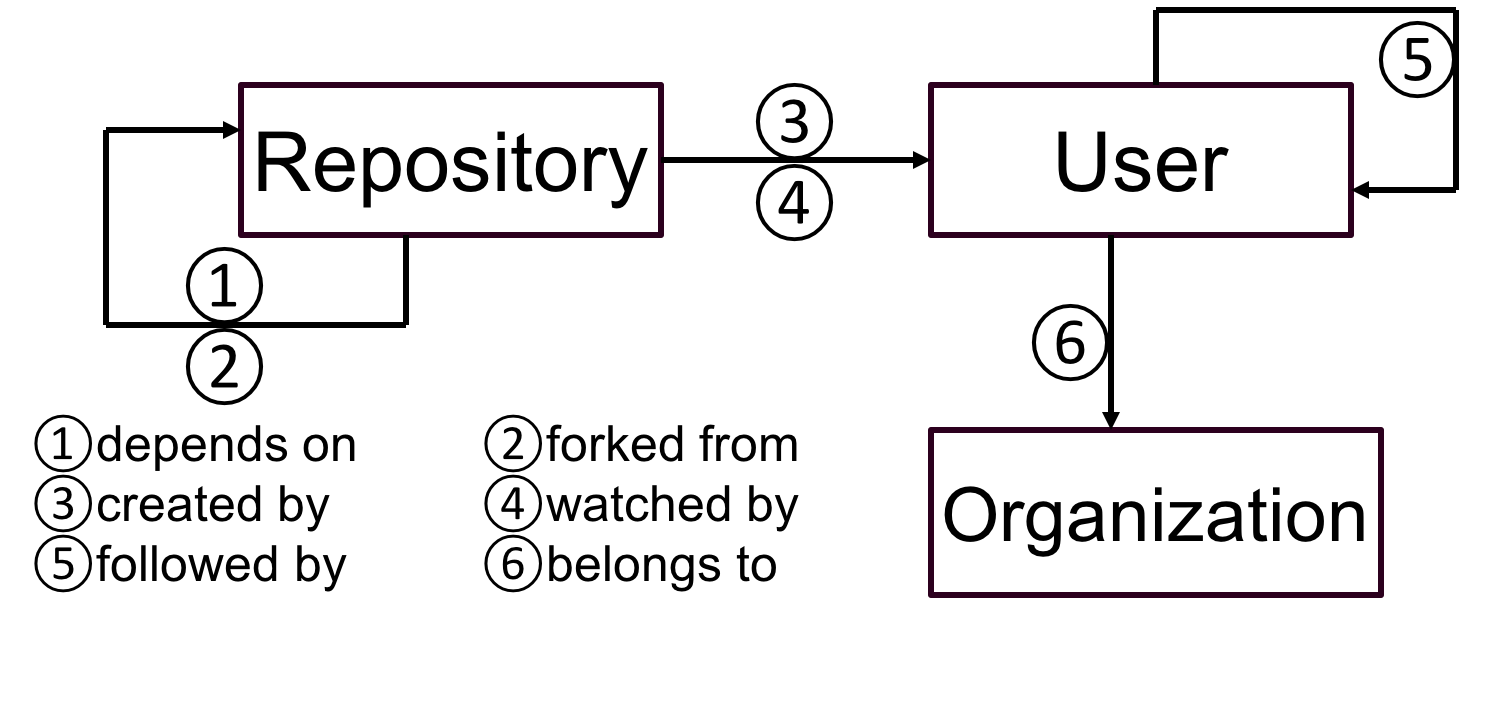}
 \caption{Schema of the GitHub heterogeneous network.}
 \label{fig:schema}
\end{figure}

Our toy example of \textsf{GitHub} data can be seen as a heterogeneous network, where the basic object types include \textsf{user}, \textsf{repository} and \textsf{organization}. The particular network schema is shown in Figure \ref{fig:schema}.

According to Definition \ref{def:hin}, in the heterogeneous network setting, due to the existence of multiple object types, labels of different types of objects can not be directly propagated, but they rather interact implicitly. For example, in our \textsf{GitHub} network in Figure \ref{fig:toy}, directly assigning the \textsf{user} label like \textsf{``ios developer''} to a \textsf{repository} object does not make much sense, but such a user label does indicate that the linked \textsf{repositories} might be more likely to be associated with labels like \textsf{``written in objective c''}. 

To capture such latent semantics and interactions of labels, as well as addressing their inherent noise and sparsity, we propose and design an object encoder to map various labels into a common embedding space  (Section \ref{sec:model}.B). As a consequence, we propagate object embeddings instead of labels on the network, and a parametric predictive model is applied to map the embeddings back to labels upon prediction. Moreover, as we will show in more details later, this object encoder can be easily extended to incorporate the rich information in various object attributes.

In heterogeneous networks, different types of objects can interact in various ways, which obviously cannot be sufficiently modeled by simple weighted links. Consider our \textsf{GitHub} network in Figure \ref{fig:toy}, where \textsf{users} can \textsf{``belong to''} \textsf{organizations} and \textsf{``create''} \textsf{repositories}. The links derived by the \textsf{``belong to''} and \textsf{``create''} relations should thus determine different label propagation functions. For example, the labels of \textsf{organizations} might be something like \textsf{``stanford university''} or \textsf{``google inc.''}, whereas those of \textsf{repositories} might be \textsf{``written in objective c''} or \textsf{``tensorflow application''}. In this case, although the labels of both \textsf{organizations} and \textsf{repositories} can influence \textsf{users}' labels regarding \textsf{``skills''} and \textsf{``interests''}, the mapping of such influences should be quite different. Moreover, consider the links even between the same types of objects, say, \textsf{users} and \textsf{repositories}. Since \textsf{users} can \textsf{``create''} or \textsf{``watch''} \textsf{repositories}, the different types of links should have different functions regarding label propagation. For example, when a \textsf{user} \textsf{``creates''} a \textsf{repository}, her labels regarding \textsf{``skills''} like \textsf{``fluent in python''} might strongly indicate the labels of the \textsf{``created''} \textsf{repository} like \textsf{``written in python''}, but when she \textsf{``watches''} a \textsf{repository}, her labels regarding \textsf{``interests''} like \textsf{``deep learning fan''} will likely indicate the labels of the \textsf{``watched''} \textsf{repository} like \textsf{``tensorflow application''}.

To model the multi-typed relations among objects, we propose to cast each type of links as a unique neural network module (Section \ref{sec:model}.C). The same module is reused over all links of the same type, so the number of parameters to be learned is independent of the size of the network, making the model efficient in memory usage and easy to train. These link-wise neural networks are jointly trained with the object encoders, so that the complex semantics in object labels (and possibly object attributes) can be well modeled to align with the various object interactions and propagation functions determined by different types of links. 

One step further, as pointed out by various existing works, we notice that the higher-order semantics in heterogeneous networks can be regulated by the tool of meta-path, defined as follows.
\begin{definition}
A \textit{meta-path} \cite{sun2012mining, sun2011pathsim} is a path defined on the network schema denoted in the form of $o_1\xrightarrow{l_1}o_2\xrightarrow{l_2}\cdots \xrightarrow{l_m}o_{m+1}$, where $o$ and $l$ are object types and link types, respectively. It represents a \textit{compositional} relation between two given object types. 
\label{def:mp}
\end{definition}

Each meta-path thus captures a particular aspect of semantics. Continue with our example on the \textsf{GitHub} network in Figure \ref{fig:toy}. The meta-path of \textsf{user}$\xrightarrow{\text{\sf creates}}$\textsf{repository}$\xrightarrow{\text{\sf watched by}}$\textsf{user} carries quite different semantics from \textsf{user}$\xrightarrow{\text{\sf belongs to}}$\textsf{organization}$\xrightarrow{\text{\sf includes}}$\textsf{user}. Thus, the two pairs of users at the ends of these two paths are similar in different ways, which are \textit{composed} by the modular links along the paths and should imply different label propagation functions.

To fully incorporate the higher-order complex semantics in heterogeneous networks, we leverage the compositional nature of paths and propose to jointly train our link-wise neural network modules through randomly sampling paths on heterogeneous networks and dynamically constructing the neural modular networks corresponding to their underlying meta-paths (Section \ref{sec:model}.D). In this way, we do not require the input of a given set of useful meta-paths, nor do we need to enumerate all legitimate ones up to a certain size. Instead, we let the random walker compose arbitrary meta-paths during training, and automatically estimate their importance and functionalities regarding LP on-the-fly. 

Finally, although NEP is powerful yet light in parameters, we deliberately designed a series of training techniques to further improve its efficiency (Section \ref{sec:model}.E). We also systematically and theoretically analyze the connections between NEP and various popular SSL algorithms, and briefly talk about several straightforward extensions of NEP left as future works (Section \ref{sec:model}.F).
  
\subsection{Object Encoder of Labels and Beyond}
Standard LP directly propagates labels on the whole network by assigning each object a label probability distribution. In this way, besides the label-object imcompatibility as we discussed before, they also ignore the complex label semantics and correlations. Moreover, labels in real-world datasets are often sparse and noisy, due to the high expense of high-quality label generation, which leads to the built-up of error rates during propagation.
 
To overcome these problems, instead of propagating labels, we propose to firstly encode various object labels into a common latent space, and then propagate the object embeddings on the network. To this end, we leverage the power of neural representation learning by jointly training an object embedding function and a label prediction function for object encoding. 

Particularly, we have the embedding $\mathbf{x}_i$ of object $v_i$ as $\mathbf{x}_i = \mathbf{f}(v_i)$.
In the simplest case, $\mathbf{f}(\cdot)$ can be implemented as a randomly initialized learnable embedding look-up table, \ie, $\mathbf{x}_i=\mathbf{E}^T \mathbf{u}_i$, where $\mathbf{E}\in \mathbb{R}^{N\times K}$ is the embeddings of the total $N$ objects on the network $\mathcal{G}$ into a $K$-dimensional latent space, and $\mathbf{u}_i$ is the one-hot vector representing the identity of $v_i$.

To encode various labels into a common latent space, we apply an MLP on the object embedding as a parametric label prediction model and impose a \textit{supervised loss} in terms of cross-entropy on softmax classification \wrt~ground-truth labels on labeled objects.
\begin{align}
\mathcal{J}_l = -\sum_{i=1}^M\log p(v_i, y_i) = -\sum_{i=1}^M \log \frac{\exp(\mathbf{W}_l ^{y_i}\mathbf{\tilde{x}}_i)}{\sum_{y\in\mathcal{Y}}\exp(\mathbf{W}_l ^y\mathbf{\tilde{x}}_i)},
\label{eq:sup}
\end{align}
where $M=|\mathcal{V}_l|$ is the number of labeled objects, $y_i$ is the ground-truth label of object $v_i$, and $\mathcal{Y}$ is the set of all distinct labels on the network. It is trivial to encode multiple labels for a single object, by computing $\sum_{y\in\mathbf{y}_i}p(v_i, y)$. 

Moreover, we have
\begin{align}
\mathbf{\tilde{x}} = \mathbf{h}^{Q_n}_n(\ldots\mathbf{h}^1_n(\mathbf{x})\ldots),
\end{align}
where
\begin{align}
\mathbf{h}^q_n(\mathbf{x}) = \text{ReLU} (\mathbf{W}_n^q\mathbf{h}_n^{q-1}(\mathbf{x})+\mathbf{b}_n^q).
\end{align}
$Q_n$ is the number of layers in the MLP, $\mathbf{W}_n^q$ and $\mathbf{b}^q$ are the parameters of the $q$-th layer, and $\mathbf{h}^0(\mathbf{x})=\mathbf{f}(\mathbf{x})$. We use $\Theta_n=\{\mathbf{W}_l, \mathbf{W}_n, \mathbf{b}_n\}$ to denote all parameters in the MLP-based parametric prediction model. As we motivated above, such an MLP is useful in capturing the complex label semantics and correlations, and at the same time address the label noise and sparsity. 

Note that, discussed above is a basic object encoder that only considers object labels. As to be shown in Section \ref{sec:model}.F, it is straightforward to extend this object encoder to consider the rich information of available attributes $\mathcal{A}$ associated with objects.

\subsection{Type-Aware Link-Wise Modules}
Now we consider the process of embedding propagation on heterogeneous networks, where multiple types of objects interact in rather complex ways. Our key insight here is, if we regard each link in the network as an influence propagation channel which allows the connected objects to influence each other, then different link types should naturally determine different propagation functions. To explicitly leverage this insight, we use a unique neural network to model the propagation functions of each type of links, which acts as a reusable module on the whole network. 

Particularly, for each module, we still resort to the MLP of feedforward neural networks, due to its compatibility with the object encoder and representation power to model the complex label-link interactions, as well as model simplicity. For each link type $t\in\mathcal{T}$, we have
\begin{align}
\mathbf{g}_t(\mathbf{x}) =  \mathbf{h}_t^{Q_t}(\ldots\mathbf{h}_t^1(\mathbf{x})\ldots),
\end{align}
where
\begin{align}
\mathbf{h}^q_t(\mathbf{x}) = \text{ReLU} (\mathbf{W}_t^q\mathbf{h}_t^{q-1}(\mathbf{x})+\mathbf{b}_t^q).
\end{align}
$\forall t\in\mathcal{T}, Q_t$ is the number of layers in the MLP, $\mathbf{W}_t^q$ and $\mathbf{b}_t^q$ are the parameters of the $q$-th layer, and $\mathbf{h}_t^0(\mathbf{x})=\mathbf{f}(\mathbf{x})$. We use $\Theta_m=\{\mathbf{W}_t, \mathbf{b}_t\}_{t\in\mathcal{T}}$ to denote all parameters in all of the MLP-based link-wise neural network modules. Note that, we use $\mathcal{T}$ to denote the set of all link types, and each link type is counted twice by considering the propagation directions.
As we will show in the experiments, compared with linear MLP, non-linear MLP allows the model of object interactions to be more flexible and effective.

Equipped with such link-wise propagation functions, to mimic the process of LP from object $v_i$ to object $v_j$ through link $e_{ij}$, we simply input the object embedding $\mathbf{x}_i$ into the neural network module corresponding to link type $t=\psi(e_{ij})$, and get $\mathbf{g}_t(\mathbf{x}_i)$ as the propagated embedding. An \textit{unsupervised loss} (\eg, an $\ell_{2}$-loss) is then computed between the propagated embedding of $v_i$ on $v_j$ and the current embedding of $v_j$ to require the \textit{label smoothness} among $v_i$ and $v_j$, \textit{conditioned} on their particular link type $\psi (e_{ij})$. Specifically, for all linked pairs of objects on the network, we have
\begin{align}
\mathcal{J}_u = \sum_{e_{ij}\in\mathcal{E}} ||\mathbf{g}_{\psi(e_{ij})}(\mathbf{x}_i) - \mathbf{x}_j||^2_2.
\label{eq:unsup}
\end{align}
Multiple links among the same pair of objects can also be trivially considered with our model by adding up all corresponding losses. 

By combining the supervised loss in Eq.~\ref{eq:sup} and unsupervised loss in Eq.~\ref{eq:unsup}, we can simply arrive at the overall loss function of NEP, which implements SSL over a heterogeneous network as follows.
\begin{align}
\mathcal{J} = \mathcal{J}_l + \lambda\mathcal{J}_u,
\end{align}
which shares the identical form with the general objective function of LP \cite{zhu2002learning, zhu2003semi} and various other SSL algorithms. By properly optimizing $\mathcal{J}$, we can jointly train our neural object encoder and link-wise modules, so that the embedding propagation along each link is jointly decided by both the end objects (particularly the propagated labels in our current model) and the link type.

\subsection{\large Comprehensive Semantics with Path Sampling}
We notice that most existing models on heterogeneous networks including the recent works on SSL \cite{jiang2017semi, chen2017task} all leverage the tool of meta-paths to capture fine-grained semantics regarding the higher-order interactions involving multiple object and link types. However, all of them explicitly model a limited set of meta-paths, which only leverages part of all complex interactions. 

In this work, we leverage the \textit{compositional nature} of paths, and propose to dynamically sample uniform random walks on heterogeneous networks and compose the corresponding modular neural networks during model training with ultimate flexibility on-the-fly. In this sense, our consideration of meta-paths is truly data-driven, \ie, the sampled paths, while naturally preferring the more common and important underlying meta-paths in particular heterogenous networks, can actually also reveal any possible meta-paths. Therefore, we are able to avoid the explicit consideration of any limited sets of meta-paths and capture the comprehensive higher-order semantics in arbitrary heterogeneous networks. 

When training NEP, instead of limiting the embedding propagation along direct links, we consider it along paths consisting of multiple links. Particularly, we revise the unsupervised loss function in Eq.~\ref{eq:unsup} into 
\begin{align}
\mathcal{J}_u' = \sum_{p_{ij}\in\mathcal{P}} ||\mathbf{G}_{p_{ij}}(\mathbf{x}_i) - \mathbf{x}_j||^2_2,
\label{eq:com}
\end{align}
where $p_{ij}$ is a path sampled with uniform random walks on the heterogeneous network, and $\mathcal{P}$ is the set of all randomly sampled paths. Correspondingly, we have $\mathcal{J}'=\mathcal{J}_l+\lambda\mathcal{J}_u'$.

\begin{definition}
A \textit{uniform random walk} in heterogeneous networks is a random walk that ignores object types. Particularly, on object $v_i$, the random walker picks the next object $v_j$ to go to based on the uniform link distribution with $p(e_{ij}|v_i) = 1/ \text{deg}(v_i)$, where deg$(\cdot)$ is the total number of all types of links that connect to $v_i$. We do not consider self-loops or restarts.
\end{definition}

Next we talk about the construction of $\mathbf{G}_{p_{ij}}$, by starting with the definition of a path $p_{ij}$.

\begin{definition}
A \textit{path} $p_{ij}$ is an ordered list $(v_i, e_1, e_2, \ldots, e_n, v_j)$, where $v_i$ and $v_j$ are the source and destination objects, respectively. $(e_1\ldots, e_n)$ are the links along the path. $n$ is the number of links in the path, and a path with $n$ links is called a length-$n$ path.
\end{definition}

With the link-wise neural network modules defined in Section \ref{sec:model}.C, we further leverage the idea of modular neural networks from visual question answering \cite{andreas2016neural, andreas2016learning}, by dynamically constructing $\mathbf{G}_{p_{ij}}$ \wrt~the underlying meta-path of $p_{ij}$ as follows.
\begin{align}
\mathbf{G}_{p_{ij}} = \mathbf{g}_{\psi(e_1)} \circ \mathbf{g}_{\psi(e_2)} \circ \cdots \circ \mathbf{g}_{\psi(e_n)}.
\label{eq:comp}
\end{align}

As we can see, by stacking the corresponding link-wise modules in the correct order, each meta-path now corresponds to a unique neural network model, where the components can be jointly trained and reused. As a consequence, each meta-path determines a unique learnable embedding propagation function, which further depends on the propagation functions of all of its component links. On one hand, the dynamically composed path-wise models capture the complex fine-grained higher-order semantics in the heterogeneous networks, while on the other hand, the learning of the link-wise modules is enhanced across the training based on various paths. As a result, NEP can be efficiently trained to deeply capture the comprehensive semantics and importance of any arbitrary meta-paths regarding the embedding propagation functions, which totally breaks free the requirements of given set of meta-paths and explicit search or learning for linear importance weights \cite{shang2016meta, jiang2017semi, luo2014hetpathmine}.

\subsection{Further Efficiency Improvements}
To further improve the efficiency of NEP, we design a series of intuitive and effective strategies.

\header{Focusing on Targeted Types of Objects.}
Our NEP framework is designed to model heterogeneous networks with multiple types of objects, which naturally can be associated with multiple sets of labels. However, in some real-world scenarios, we only care about the labels of some particular \textit{targeted} types of objects. For example, when we aim to classify \textsf{repositories} on \textsf{GitHub}, we are not explicitly interested in \textsf{user} and \textsf{organization} labels. 

Due to this observation, we can aggressively simplify NEP by only computing the embeddings of targeted types of objects and subsequently constraining the random paths to be sampled only among them. We call this model NEP-target. Compared with NEP-basic, NEP-target allows us to significantly reduce the size of the embedding look-up tables in the object encoder by ${\it 35\%-65\%}$, which costs the most memory consumption, compared with other model parameters that are irrelevant to the network sizes. Moreover, since we focus on the embedding propagation among targeted types of objects, NEP-target can effectively save the time of learning the encodings of non-targeted types of objects, which leads to about ${\it 60\%}$ \textit{shorter runtimes} until convergence compared with NEP-basic. Finally, it also helps to alleviate the built-up of noises and errors when propagating through multiple poorly encoded intermediate objects, which results in ${\it 12\%-21\%}$ \textit{relative performance gain} compared with NEP-basic. 

Note that, ignoring the embedding of non-targeted objects does not actually contradict with our model motivation, which is to capture the complex interactions among different types of objects. 
This simplification only works in particular scenarios like the ones we consider in this work, where we only care about and have access to the labels of particular types of objects, and the identities of non-targeted objects are less useful without the consideration of their labels and attributes. In this case, the only information that matters for the non-targeted types of objects is their types, which is sufficiently captured by our type-aware link-wise modules.

\header{Fully Leveraging Labeled Data.}
By focusing on targeted types of objects, we have saved a lot of training time for learning the embeddings of non-targeted types of objects. However, on real-world large-scale networks, learning the embeddings of unlabeled targeted types of objects can still be rather inefficient. This is because the embeddings of most objects (\ie, unlabeled objects) are meaningless at the beginning, and therefore the modeling of their interactions is also wasteful. 

Our first insight here is, to fully leverage labeled data, we should focus on paths that include at least one labeled object, whose embedding directly encodes label information. Since our modular neural networks are reused everywhere in the network, the propagation functions of different links and paths captured around labeled objects are automatically applied to those among unlabeled objects. Moreover, due to the small diameter property of real-world networks \cite{watts1998collective}, we assume that moderately long (\eg, length-4) paths with at least one labeled object can reach most unlabeled objects for proper learning of their embeddings.

One step further, we find it useful to only focus on paths ending on labeled objects. The insight here is, according to Eq.~\ref{eq:com}, the $\ell_{2}$-loss is computed between the propagated embedding $\mathbf{G}_{p_{ij}}(\mathbf{x}_i)$ of the start object $v_i$ and the current embedding $\mathbf{x}_j$ of the end object $v_j$, so training is more efficient if at least one of the two embeddings is ``clean'' by directly encoding the label information. In this case, $\mathbf{x}_j$ is clean if $v_j$ is labeled, but $\mathbf{G}_{p_{ij}}(\mathbf{x}_i)$ is not clean even if $v_i$ is labeled. Therefore, we apply \textit{reverse} path sampling, \ie, we always sample paths from labeled objects, and use them in the reverse way, to make sure the end objects are always labeled.

We call this further improved model variant NEP-label.
In our experiments, we observe that NEP-label leads to another ${\it 85\%-96\%}$ \textit{shorter runtimes} until convergence and ${\it 2.3\%-26\%}$ \textit{relative performance} gain compared with NEP-target.

\header{Training with Two-Step Path Sampling,}
As we have discussed in Section \ref{sec:model}.D, one major advantage of NEP over existing SSL models on heterogeneous networks is the flexibility of considering arbitrary meta-paths and training the corresponding modular networks with path sampling based on uniform random walks on-the-fly, by leveraging the compositional nature of paths. However, since the modular networks composed for different paths have different neural architectures, it poses unique challenges for the efficient training of NEP by leveraging batch training, especially on GPUs.

We notice that, according to Eq.~\ref{eq:comp}, paths sharing the same underlying meta-path should correspond to the same composed modular network. Therefore, although paths sampled by uniform random walks can be arbitrary, we can always group them into smaller batches according to their underlying meta-paths. However, path grouping itself is time consuming, and it leads to different group sizes, which is still not ideal for efficient batch training.

To address the challenges, we design a novel two-step path sampling approach for the efficient training of NEP, as depicted in Algorithm 1. Specifically, in order to sample a total number of $\Omega$ paths (\eg, 100K), we firstly sample a smaller set of $\Gamma$ paths (\eg, 100). Then for each of these $\Gamma$ paths, we find its underlying meta-path $\mathcal{M}$, and sample $B=\Omega/\Gamma$ paths (\eg, 1K) that follow $\mathcal{M}$. Therefore, the particular modular network corresponding to $\mathcal{M}$ can be composed only once and efficiently trained with standard gradient back-propagation with the batch of $B$ samples. 

To sample random paths guided by particular meta-paths in Step 13, we follow the standard way in \cite{dong2017metapath2vec, shang2016meta}. However, different from them, our meta-paths are also sampled from the particular network, rather than given by domain experts or exhaustively enumerated. Assuming $\Omega$ is sufficiently large compared with $B$, the total $\Omega$ paths sampled by our two-step approach only differ in orders from any $\Omega$ paths sampled by the original approach, which corresponds to the special case with $B=1$. Therefore, our path sampling approach is purely data-driven, and our model automatically learns their importance and complex functions regarding embedding propagation.

In Section \ref{sec:exp}, we show that our two-step path sampling approach can significantly reduce the runtimes of NEP, while it is also able to slightly boost its performance, due to more stable training and faster convergence. 

\header{Training Algorithm.} 
Algorithm 1 gives an outline of our overall training process, which is based on NEP-label with two-step path sampling. In the inner loop starting from Line 6, it samples a path completely at random without the consideration of meta-path. Then it samples more instances under the same meta-paths. This strategy is crucial to eliminate the explicit consideration of a limited set of meta-paths, which makes NEP different from all existing SSL algorithms on heterogeneous networks. It is also crucial for leveraging batch-wise gradient backpropagation, which utilizes the power of dynamically composed modular networks. In this way, our model is data driven and able to consider any possible meta-paths underlying uniform random walks on heterogeneous networks. 

\begin{algorithm}[h!]
\caption{Efficient Training of NEP}
\begin{algorithmic}[1]
\Procedure{NEPTrain}{}
	\State \textbf{Input:} $\mathcal{G}$, $\Omega$, $\Gamma$, $B (=\Omega/\Gamma)$, max path length $L$
	\For{$i\gets 1$ to $\Gamma$}
		\State Sample a source object $v_s\in \mathcal{V}_l$
		\State $p=(v_s)$
		\For{$j\gets 1$ to $L$}
			\State Sample $(e_{ij}, v_j)$ on $\mathcal{G}$ from $p[-1]$ and append to $p$ 
			\If {$\phi(v_j) \in $targeted object type}
				\State break
			\EndIf
		\EndFor
		\State Find the underlying meta-path $\mathcal{M}$ of $p$
		\State Sample $B$ paths under $\mathcal{M}$ from random labeled objects
		\State Take a gradient step to optimize $\mathcal{J}'$
	\EndFor
\EndProcedure
\end{algorithmic}
\end{algorithm}

\header{Complexity Analysis.}
In terms of memory, the number of parameters in NEP is $O(N+L+T)$, where $N$ is the number of targeted types of objects in the network, $L$ and $T$ are the number of classes and number of link types, respectively, which are independent of the network sizes. The $O(N)$ term is due to the embedding look-up table $\mathbf{E}$, which can be further reduced to a constant number if we replace it with an MLP given available object attributes. The $O(L)$ and $O(T)$ terms are due to the parameters in $\Theta_n$ of the predictive model and $\Theta_m$ in the link-wise neural network modules, respectively. 

In terms of runtime, training NEP theoretically takes $O(\Omega L)$ time, where $\Omega$ is the number of sampled paths and $L$ is the path length. It is the same as the state-of-the-art unsupervised heterogeneous network embedding algorithms \cite{shang2016meta, dong2017metapath2vec}, while can be largely improved in practice based on the series of strategies we develop here. 

\subsection{Connections and Extensions}
We show that NEP is a principled and powerful SSL framework by studying its connections to various existing graph-based SSL algorithms and promising extensions towards further improvements on modeling heterogeneous networks.

To better understand the mechanism of NEP, we analyze it in the well-studied context of graph signal processing \cite{shuman2013emerging, girault2014semi}.
Specifically, we decompose NEP into three major components: \textit{embedding}, \textit{propagation}, and \textit{prediction}, which can be mathematically formulated as $\mathcal{X} = \mathcal{F}(\mathcal{V})$, $\mathcal{X}' = \mathcal{G}(\mathcal{X})$, and $\mathcal{\hat{Y}} =\mathcal{Z}(\mathcal{X}')$,
where $\mathcal{X}$ is the graph embedding with $\mathcal{F}$ as the embedding function, $\mathcal{X}'$ is the propagated embedding with $\mathcal{G}$ as the propagation function, and $\mathcal{\hat{Y}}$ is the label prediction with $\mathcal{Z}$ as the prediction function.

In NEP, we apply an embedding look-up table as $\mathcal{F}$ to directly capture the training labels $\mathcal{Y}$ on $\mathcal{V}_l$, while it is straightforward to implement $\mathcal{F}$ as an MLP to also incorporate object attributes $\mathcal{A}$ on $\mathcal{V}$, which we leave as future works. Such embedding allows us to further explore the complex interactions of labels and attributes among different types of objects. Our major technical contribution is then on the propagation function $\mathcal{G}$, which leverages modular networks to properly propagate object embeddings on different paths. Finally, due to the appropriate embedding and propagation functions, we are able to jointly learn a powerful parametric prediction function $\mathcal{Z}$ as an MLP based on very few samples.

\begin{table}[h!]
\centering\begin{tabular}{|c|c|c|c|}
\hline
{\bf Algorithm}& $\mathcal{F}(\mathcal{V})$ & $\mathcal{G}(\mathcal{X})$ & $\mathcal{Z}(\mathcal{X}')$ \\
  \hline
{\bf LP \cite{zhu2002learning} }& $\mathcal{Y}$ & $(I+\alpha L)^{-1}\mathcal{X}$ & $\text{argmax}_j\mathcal{X}'_{ij}$ \\
  \hline
{\bf MR \cite{belkin2006manifold} }& $\mathcal{A}$ & $(I+\alpha L)^{-1}\mathcal{X}$ & $\mathbf{W}\mathcal{X}'+\mathbf{b}$ \\
\hline
{\bf Planetoid \cite{yang2016revisiting} }& $\text{MLP}(\mathcal{A})$ & $(I+\alpha L)^{-1}\mathcal{X}^*$ & $\text{MLP}(\mathcal{X}')$ \\
\hline
{\bf GCN \cite{kipf2016semi} }& $\text{MLP}(\mathcal{A})$ & $(I - \tilde{L})^{k}\mathcal{X}$ & $\text{MLP}(\mathcal{X}')$ \\
\hline
 \end{tabular}
 \caption{ \label{tab:connection}\textbf{\small Decomposition of popular graph-based SSL algorithms.}}
 \vspace{-10pt}
\end{table}

In fact, we find that various existing graph-based SSL algorithms can well fit into this three-component paradigm, and they naturally boil down to certain special cases of NEP.
As summarized in Table \ref{tab:connection}, the classic LP algorithm \cite{zhu2002learning} directly propagates object labels on the graph based on a deterministic auto-regressive function $(I+\alpha L)^{-1}$, where $I$ is the identical matrix and $L$ is the graph Laplacian matrix \cite{spielman2007spectral}. Prediction of LP is done by picking out the propagated labels with largest values. To jointly leverage object attributes and the underlying object links, 
MR \cite{belkin2006manifold} trains a parametric prediction model based on SVM with a graph Laplacian regularizer. To some extend, it can be viewed as propagating the object attributes on the graph. 
The recently proposed graph neural network models like Planetoid \cite{yang2016revisiting} and GCN \cite{kipf2016semi} 
leverage object embedding to integrate object attributes, links and labels. Although the two works adopt quite distinct models, they essentially only differ in the implementations of the propagation function $\mathcal{G}$: 
Planetoid leverages random path sampling on networks to approximate the effect of graph Laplacian \cite{yang2017bridging}, whereas GCN sums up the embedding of neighboring objects in each of its $k$ convolutional layers through a different smoothing function with $\tilde{L}$ as the normalized graph Laplacian with self-loops.

In concept, due to the expressiveness of neural networks, NEP can learn arbitrary propagation functions given proper training data, thus generalizing all of the above discussed algorithms. 
Moreover, since we notice that the propagation functions of most SSL algorithms on homogeneous networks are constructed to act as a deterministic low-pass filter that effectively encourages smoothness among neighboring objects, it is interesting to extend NEP by designing proper constraints on our modular networks to further construct learnable low-pass filters on heterogeneous networks.

\section{Experiments}
\label{sec:exp}
In this section, we comprehensively evaluate the performance of NEP for SSL over three massive real-world heterogeneous networks. The implementation of NEP is all public on GitHub\footnote{https://github.com/JieyuZ2/NEP}.
\subsection{Experimental Settings}
\header{Datasets.}
We describe the datasets we use for our experiments as follows with their statistics summarized in Table \ref{tab:stat}.
\begin{enumerate}[leftmargin=23pt]
\item \header{DBLP}
We use the public Arnetminer dataset V8 collected by \cite{tang2008arnetminer}. It contains four types of objects, \ie, authors (A), papers (P), venues (V), and years (Y).
The link types include authors writing papers, papers citing papers, papers published in venues, and papers published in years.
\item \header{YAGO}
We use the public knowledge graph derived from Wikipedia, WordNet, and GeoNames \cite{suchanek2007yago}. There are seven types of objects in the network: person (P), location (L), organization (O), and \etc, as well as twenty-four types of links.
\item \header{GitHub} We use an anonymous social network dataset derived from the GitHub community by DARPA. It contains three types of objects: repository (R), user (U) and organization (O), and six types of links, as depicted in Figure \ref{fig:schema}.
\end{enumerate}

\begin{table}[h!]
\centering\begin{tabular}{|c|cccc|}
\hline
{\bf Dataset}&{\bf \#object}&{\bf \#edge}&{\bf \#class}&{\bf \%labeled}\\
  \hline  
  \hline
{\bf DBLP}& 4,925,160 & 44,931,742 & 4 & $0.081\%$\\
\hline
{\bf YAGO}& 545,792 & 3,517,663 & 15 & $0.166\%$ \\
\hline
{\bf GitHub}& 2,078,030 & 61,332,330 & 16 & $0.197\%$ \\
\hline
\hline
{\bf sub-DBLP}& 333,160 & 2,620,736 & 4 & $1.204\%$\\
\hline
{\bf sub-YAGO}& 15,672 & 3,57,312 & 15 & $37.219\%$\\
\hline
{\bf sub-GitHub}& 32,792 & 347,768 & 16 & $12.503\%$\\
\hline
 \end{tabular}
 \caption{ \label{tab:stat}\textbf{The statistics of datasets.}}
\end{table}

In order to compare with some state-of-art graph SSL algorithms that cannot scale up to large networks with millions of objects, we create a smaller sub-graph on each dataset by only keeping the labeled objects (both training and testing labels) and their direct neighbors. We also summarize their statistics in Table \ref{tab:stat}.

\header{Compared Algorithms.}
We compare NEP with the following graph-based SSL algorithms and network embedding algorithms:\\
$\bullet$ \textbf{LP} \cite{zhu2002learning}: Classic graph-based SSL algorithm that propagates labels on homogeneous networks. To run LP on heterogeneous networks, we suppress the type information of all objects. \\
$\bullet$ \textbf{GHE} \cite{chen2017task}: The state-of-the-art SSL algorithm on heterogeneous networks through path augmented and task guided embedding.\\
$\bullet$ \textbf{SemiHIN} \cite{jiang2017semi}: Another recent SSL algorithm with promising results on heterogeneous networks by ensemble of meta-graph guided random walks.\\
$\bullet$ \textbf{ZooBP} \cite{eswaran2017zoobp}: Another recent SSL algorithm on heterogeneous networks by performing fast belief propagation.\\
$\bullet$ \textbf{Metapath2vec} \cite{dong2017metapath2vec}: The state-of-the-art heterogeneous network embedding algorithm through heterogeneous random walks and negative sampling.\\
$\bullet$ \textbf{ESim} \cite{shang2016meta}: Another recent heterogeneous network embedding algorithm with promising results through meta-path guided path sampling and noise-contrastive estimation. \\
$\bullet$ \textbf{Hin2vec} \cite{fu2017hin2vec}: Another recent heterogeneous network embedding algorithm that exploits different types of links among nodes.

\header{Evaluation Protocols.}
We study the efficacy of all algorithms on the standard task of semi-supervised node classification. The labels are semantic classes of objects not directly captured by the networks. 
For DBLP, we use the manual labels of authors from four research areas, \ie, database, data mining, machine learning and information retrieval provided by \cite{sun2011pathsim}. For YAGO, we extract top 15 locations by the edge ``wasBornIn" as labels for the person objects and remove all ``wasBornIn" links from the network. 
For GitHub, we manually select 16 high-quality tags of repositories as labels, such as security, machine learning, and database. 

We randomly select $20\%$ of labeled objects as testing data and evaluate all algorithms on them. For the SSL algorithms, (\ie, LP, GHE, SemiHIN, ZooBP and NEP), we provide the rest 80\% labeled objects as training data. 
For the unsupervised embedding algorithms, (\ie, Metapath2vec, ESim and Hin2vec), we compute the embedding without training data.
For all algorithms that outputs a network embedding (\ie, Metapath2vec, ESim, Hin2vec and NEP), we train a subsequent MLP with the same architecture on the learned embeddings with the same 80\% labeled training data to predict the object classes. 
Besides the standard classification accuracy, we also record the runtimes of all algorithms which are measured on a server with four GeForce GTX 1080 GPUs and a 12-core 2.2GHz CPU. For the sake of fairness, we run all algorithms with a single thread.

\header{Parameter Settings.}
For all datasets, we set the batch size $B$ to 1,000 and learning rate to 0.001. 
For different datasets, the number of sampled patterns $\Gamma$ and maximum path length $L$ are set differently as summarized in Table \ref{tab:para}. 
For all embedding algorithms, we set the embedding dimension to 128 for full graphs and 64 for sub-graphs. Other parameters of the baseline algorithms are set as the default values as suggested in the original works.
For algorithms that require given sets of meta-paths (\ie, GHE, SemiHin, Metapath2vec and ESim), since the schemas of our experimented heterogeneous networks are relatively simple, we compose and give them the commonly used meta-paths\footnote{DBLP: A-P-A, A-P-P-A, A-P-A-P-A, A-P-V-P-A, A-P-Y-P-A; YAGO:P-P, P-W-P, P-R-P, P-S-P, P-O-P, P-D-P, P-D-D-P, P-D-E-D-P; GitHub: R-R, R-U-R, R-U-U-R, R-U-R-U-R, R-U-O-U-R. The weights are all uniform.}. 
For NEP, we use single-layer MLPs (one fully-connected layer plus one Sigmoid activation layer) for all modules with the same size. 
We have also done a comprehensive study of the impacts of major hyper-parameters.

\begin{table}[h]
\centering
   \resizebox{0.48\textwidth}{!}{\begin{tabular}{|c|c|c|c|c|c|c|}
   \hline
{\bf Dataset}&{\bf DBLP}&{\bf YAGO}&{\bf GitHub}&{\bf sub-DBLP}&{\bf sub-YAGO}&{\bf sub-GitHub}\\
  \hline
{\bf $\Gamma$}& 9000 & 7000 &6000 &6000 &2000 & 2000\\
\hline
{\bf $L$}& 5 & 6 & 6 & 7 & 5 & 4\\
\hline
 \end{tabular}}
 \caption{ \label{tab:para}\textbf{The number of sampled patterns and maximum path lengths for NEP on different datasets.}}
\end{table}

\header{Research Problems.} Our experiments are designed to answer the following research questions:
\begin{itemize}[leftmargin=15pt]
\item  \header{Q1. Effectiveness} Given limited labeled data, how much does NEP improve over the state-of-the-art graph-based SSL and network embedding algorithms?
\item  \header{Q2. Efficiency} How efficient is NEP regarding the leverage of labeled data and computational resources? 
\item  \header{Q3. Robustness} How robust is NEP regarding different settings of model hyper-parameters?
\end{itemize}

\subsection{Q1. Effectiveness}
We quantitatively evaluate NEP against all baselines on the standard node classification task. Table \ref{tab:overall} shows the performance of all algorithms on the six datasets. All algorithms are trained and tested with 10 runs on different randomly split labeled data to compute the average classification accuracy. The performance gain of NEP over baselines all passed the significant test with $p$-value 0.005. The performance of baselines varies across different datasets, while NEP is able to constantly outperform all of them with significant margins, demonstrating its supreme and general advantages. 

\begin{table}[h!]
\small
\centering
  \resizebox{0.48\textwidth}{!}{\begin{tabular}{|c|c|c|c|}
   \hline
{\bf Algorithm}&{\bf sub-DBLP}&{\bf sub-YAGO}&{\bf sub-GitHub}\\
\hline
{\bf LP}& 0.783$\pm$0.000 & 0.597$\pm$0.008 & 0.374$\pm$0.003\\
\hline
{\bf GHE}& 0.778$\pm$0.014 & 0.501$\pm$0.002 & 0.353$\pm$0.011\\
\hline
{\bf SemiHIN}& 0.787$\pm$0.000 & 0.630$\pm$0.000 & 0.306$\pm$0.000\\
\hline
{\bf ZooBP}& 0.680$\pm$0.000 & 0.382$\pm$0.000 & 0.312$\pm$0.000\\
\hline
{\bf Metapath2vec}& 0.851$\pm$0.003 & 0.604$\pm$0.003 & 0.384$\pm$0.003\\
\hline
{\bf ESim}& 0.824$\pm$0.005 & 0.563$\pm$0.004 & 0.342$\pm$0.006\\
\hline
{\bf Hin2vec}& 0.856$\pm$0.005 & 0.628$\pm$0.005 & 0.341$\pm$0.003\\
\hline
{\bf NEP-linear}& {0.885$\pm$0.003} & {0.648$\pm$0.003} & {0.400$\pm$0.180}\\
\hline
{\bf NEP}& \bf{0.888$\pm$0.005} & \bf{0.651$\pm$0.002} & \bf{0.425$\pm$0.007}\\
\hline
\hline
{\bf Algorithm}&{\bf DBLP}&{\bf YAGO}&{\bf GitHub}\\
\hline
{\bf LP}& 0.811$\pm$0.033 & 0.612$\pm$0.004 & 0.340$\pm$0.006\\
\hline
{\bf GHE}& 0.759$\pm$0.048 & 0.447$\pm$0.020 & 0.351$\pm$0.019\\
\hline
{\bf SemiHIN}& 0.724$\pm$0.000 & 0.457$\pm$0.000 & 0.348$\pm$0.000\\
\hline
{\bf ZooBP}& 0.610$\pm$0.000 & 0.561$\pm$0.000 & 0.302$\pm$0.000\\
\hline
{\bf Metapath2vec}& 0.790$\pm$0.005 & 0.590$\pm$0.005 & 0.320$\pm$0.007\\
\hline
{\bf ESim}&0.647$\pm$0.010 & 0.607$\pm$0.004 & 0.305$\pm$0.003\\
\hline
{\bf Hin2vec}&0.836$\pm$0.001 & 0.609$\pm$0.004 & 0.338$\pm$0.006\\
\hline
{\bf NEP-linear}& 0.865$\pm$0.003 & 0.629$\pm$0.002 & 0.384$\pm$0.002\\
\hline
{\bf NEP}& \bf{0.880$\pm$0.006} & \bf{0.634$\pm$0.005} & \bf{0.392$\pm$0.011}\\
\hline
 \end{tabular}}
 \caption{ \label{tab:overall}\textbf{Overall effectiveness of compared algorithms.}}
\end{table}

Taking a closer look at the scores, we observe that NEP is much better than baselines on the sub-graphs of YAGO and GitHub, where the graphs have relatively complex links but small sizes. For example, in GitHub, a repository and a user can have ``watched by'' and ``created by'' links, while in YAGO, a person and a location can have ``lives in'', ``died in'', ``is citizen of'' and other types of links. On these graphs, NEP easily benefits from its capability of distinguishing and leveraging different types of direct interactions through the individual neural network modules. 
However, when evaluated on full graphs, the performance gain of NEP is larger on DBLP. The fact is, since direct interactions are simpler in DBLP, where only a single type of link exists between any pair of objects, higher-order interactions matter more. For example, a path of ``A-P-V-P-A'' exactly captures the pairs of authors within the same research communities. The full graphs, compared with the sub-graphs, can provide much more instances of such longer paths, and NEP effectively captures such particular higher-order interactions through learning the dynamically composed modular networks. 

The advantage of NEP mainly roots in two perspectives: the capability of modeling the compositional nature of meta-paths and the flexibility of non-linear propagation functions. To verify this, we also implement a linear version of NEP by simply removing all non-linear activation functions. As we can clearly see, the performance slightly drops after removing the nonlinearity in a consistent way.

\subsection{Q2. Efficiency}
In this subsection, we study the efficiency of NEP regarding the leverage of both labeled data and computational resources.

One of the major motivations of NEP is to leverage limited labeled data, so as to alleviate the deficiency of deep learning models when training data are hard to get. Therefore, we are interested to see how NEP performs when different amounts of training data are available. 
To this end, we change the amount of training data from $10\%$ to $90\%$ of all labeled data, while the rest $10\%$ labeled data are held out as testing data. We repeat the same process but split the data randomly for 10 times to take the average scores.  

\begin{figure*}[h]
\centering
\subfigure{
\includegraphics[width=0.29\textwidth]{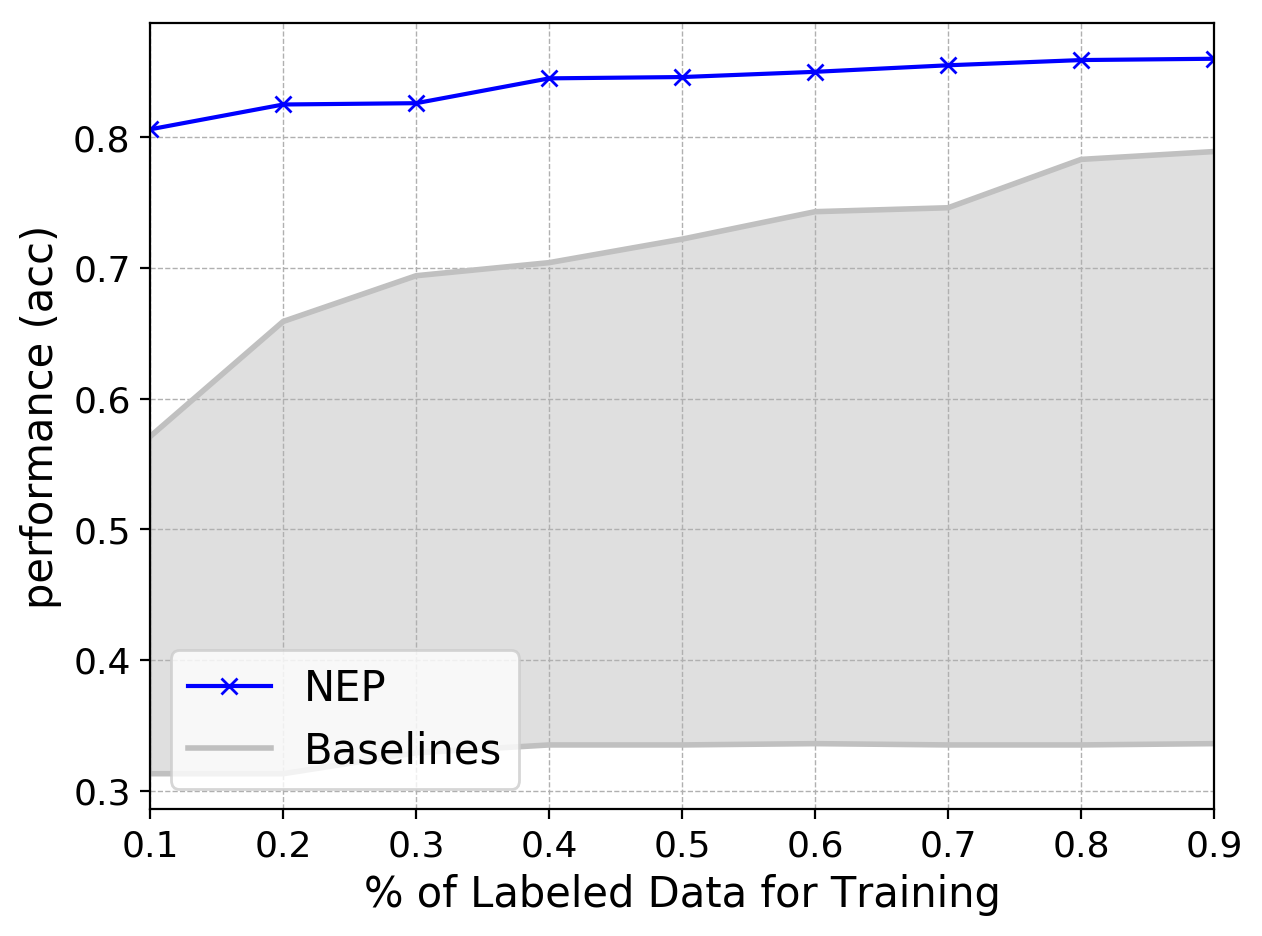}}
\subfigure{
\includegraphics[width=0.29\textwidth]{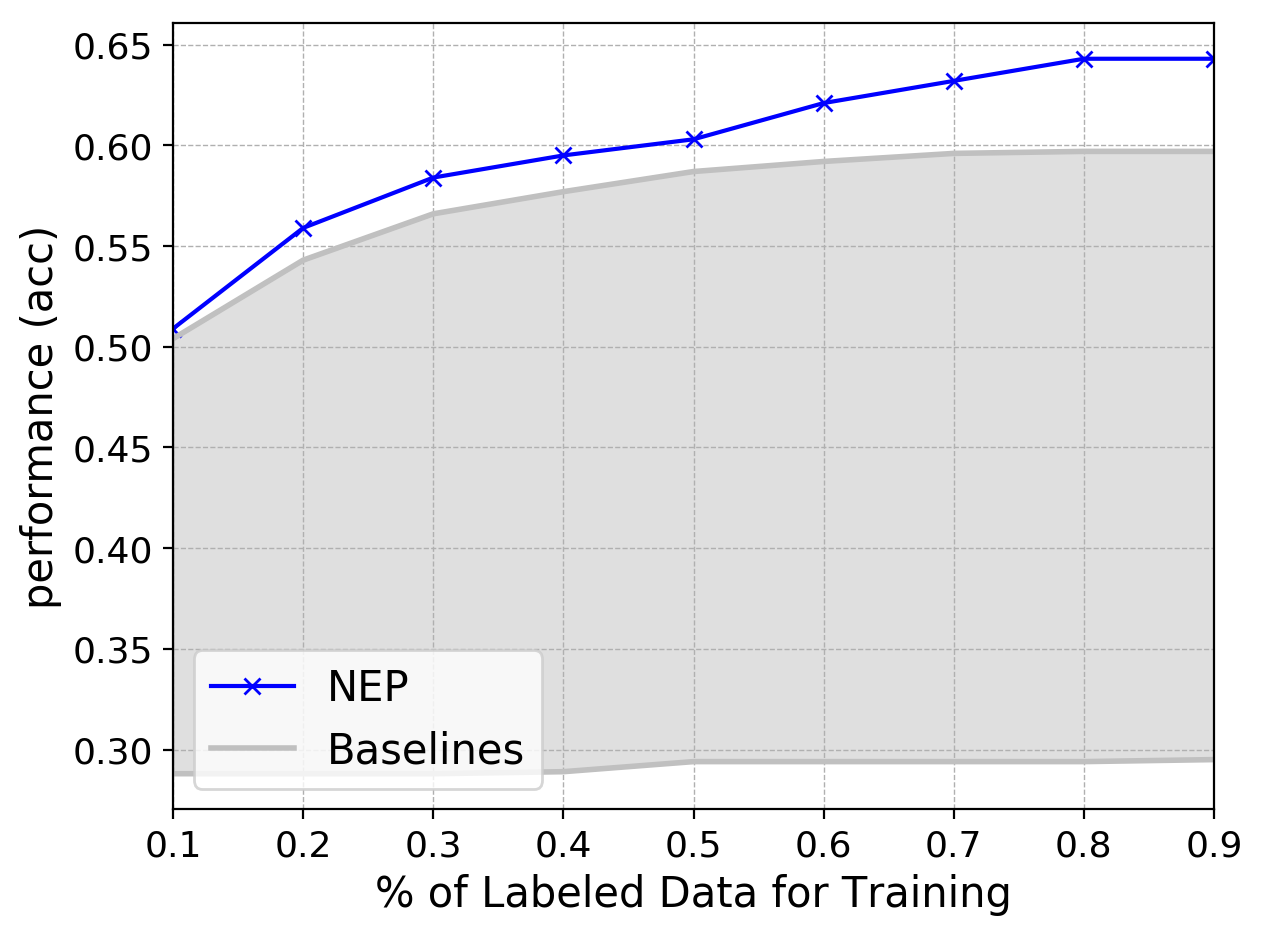}}
\subfigure{
\includegraphics[width=0.29\textwidth]{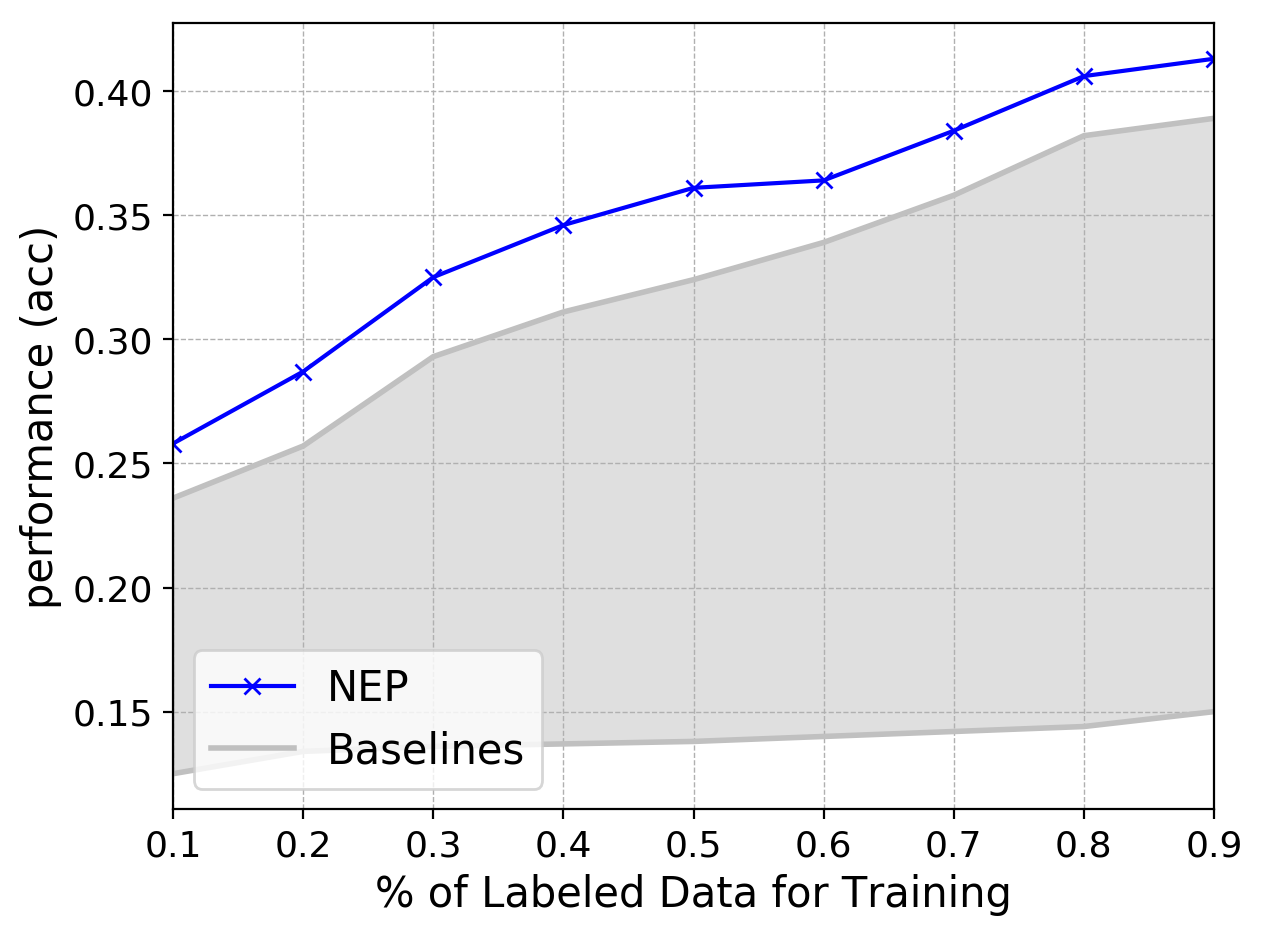}}
\vspace{-5pt}
\caption{\textbf{Efficiency on leveraging limited labeled data (from left to right: DBLP, YAGO, GitHub).}}
\label{fig:label}
\end{figure*}

As we can observe in Figure \ref{fig:label}, NEP can quickly capture the simple semantics in DBLP and reaches stable performance given only $10\%-20\%$ of the labeled data. Although YAGO and GitHub appear to be more complex and require relatively more training data, NEP maintains the best performances compared with all baselines. 
Such results clearly demonstrates the efficiency of NEP in leveraging limited labeled data.

Another major advantage of neural network models is that they can usually be efficiently trained on powerful computation resources like GPUs with well-developed optimization methods like batch-wise gradient backpropagation. 
Particularly for NEP, as we have discussed in Section \ref{sec:model}.E, since our modular networks are dynamically composed according to randomly sampled paths, we have developed a novel training strategy based on two-step path sampling to fully leverage the computational resources and standard optimization methods. Here we closely study its effectiveness.

\begin{figure*}[h]
\centering
\subfigure{
\includegraphics[width=0.29\textwidth]{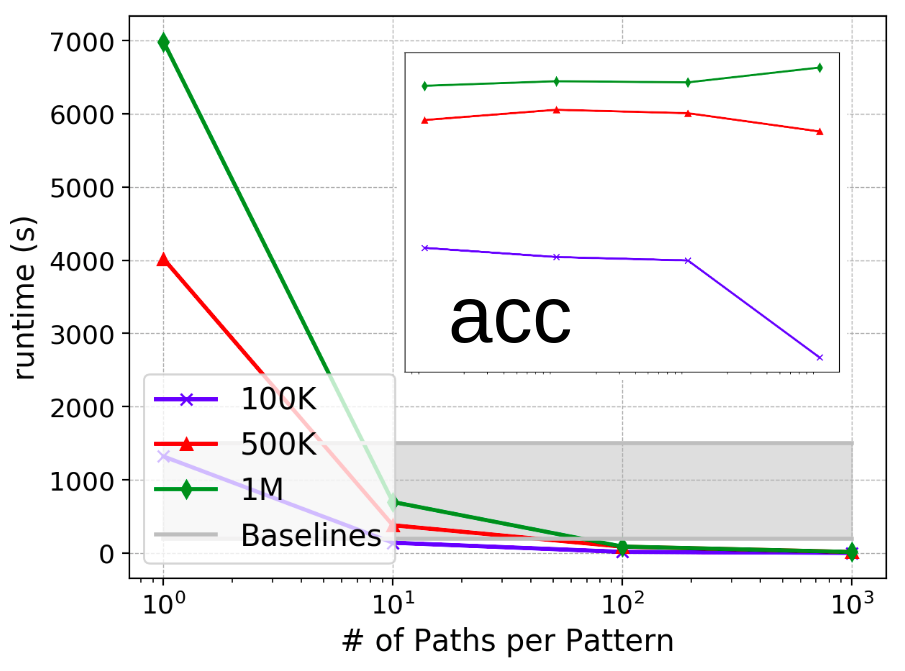}}
\subfigure{
\includegraphics[width=0.29\textwidth]{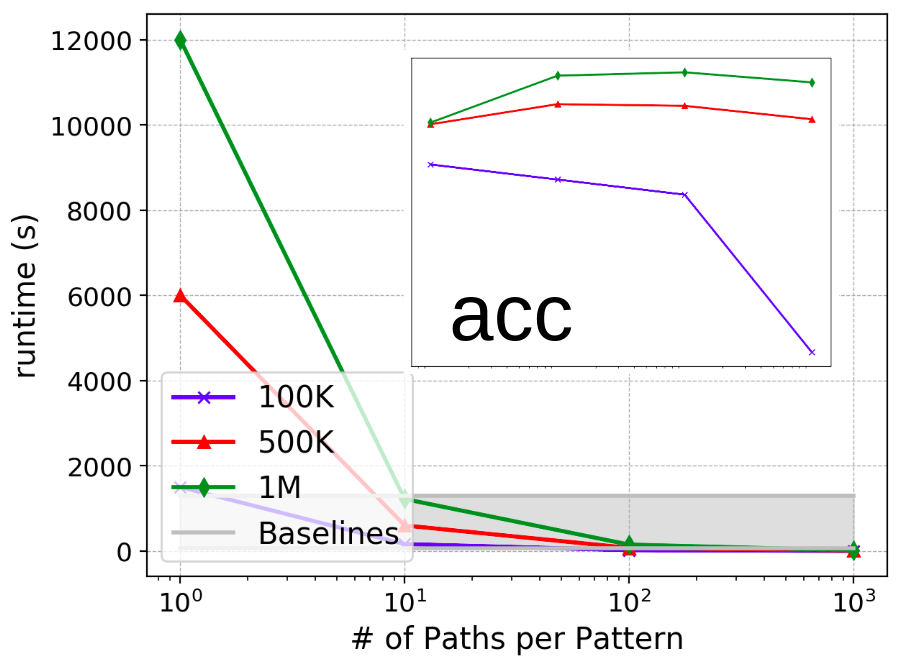}}
\subfigure{
\includegraphics[width=0.29\textwidth]{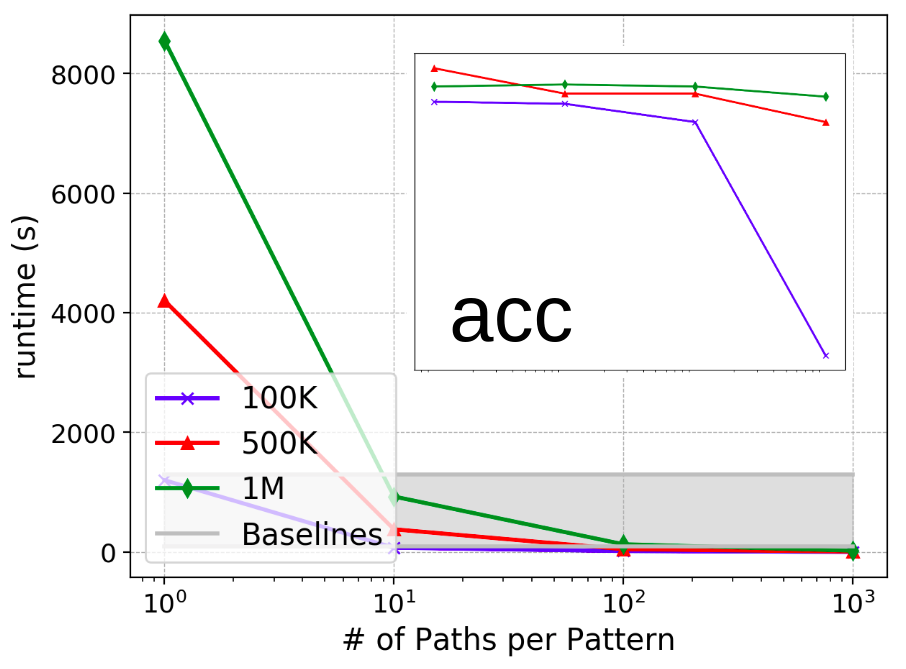}}
\vspace{-5pt}
\caption{\textbf{Efficiency on leveraging computational resources (from left to right: DBLP, YAGO, GitHub).}}
\label{fig:compute}
\end{figure*}

Figure \ref{fig:compute} shows how the strategy of training with two-step path sampling influences the performance and runtime of NEP regarding different settings on the three datasets.
To present a comprehensive study, we set the total number of sampled paths ($\Omega$) to 100K, 500K, and 1M, respectively, and then simultaneously vary the number of patterns ($\Gamma$) and the number of paths per pattern ($B$). 
As we can observe from the results, when $B=1$, which equals to no usage of two-step sampling, the runtimes are quite high; as we increase $B$ to $10$ and $10^2$, the runtimes rapidly drop, while the performances are not influenced much. Sometimes the performance actually increases, probably due to better convergence of the loss with batch training. Setting $B$ to too large values like $10^3$ does hurt the performance, but in practice, we can safely avoid it by simply setting $B$ to an appropriate value like $10^2$, which leads to a satisfactory trade-off between effectiveness and efficiency across different datasets.

\subsection{Q3. Robustness}
We comprehensively study the robustness of NEP regarding different hyper-parameter settings. 

We firstly look at the path length $L$. As shown in Table \ref{tab:length}, the performance of NEP regarding different path lengths does not differ significantly as we vary $L$ from 2 to 7. This is because shorter paths are usually more useful, and when $L$ is large, Algorithm 1 often automatically stops the sampling process at Line 9 upon reaching targeted types of objects before the actual path length reaches $L$. Therefore, in practice, the rule-of-thumb is to simply set $L$ to larger values like 5-6.

\begin{figure*}[h]
\centering
\subfigure{
\includegraphics[width=0.29\textwidth]{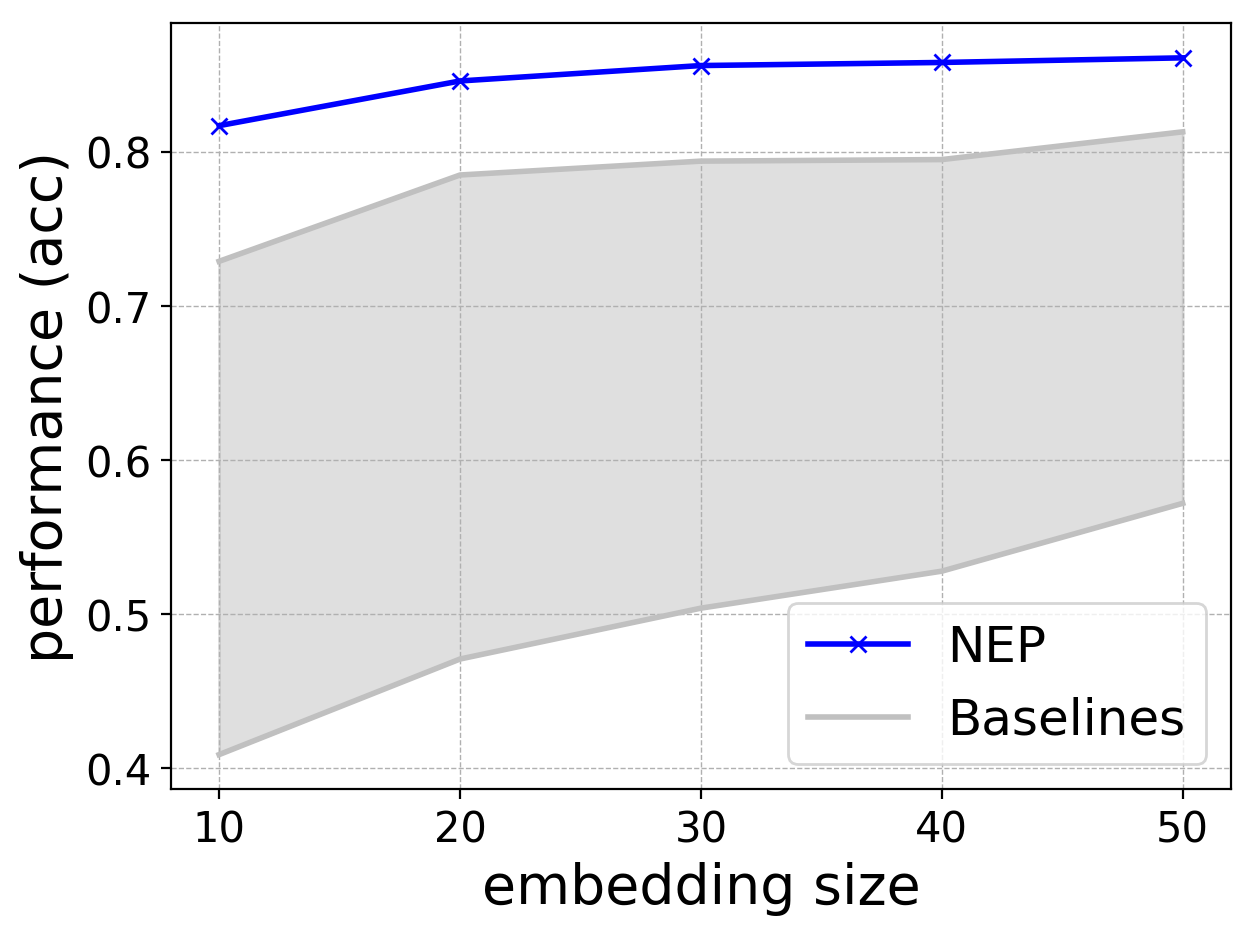}}
\subfigure{
\includegraphics[width=0.29\textwidth]{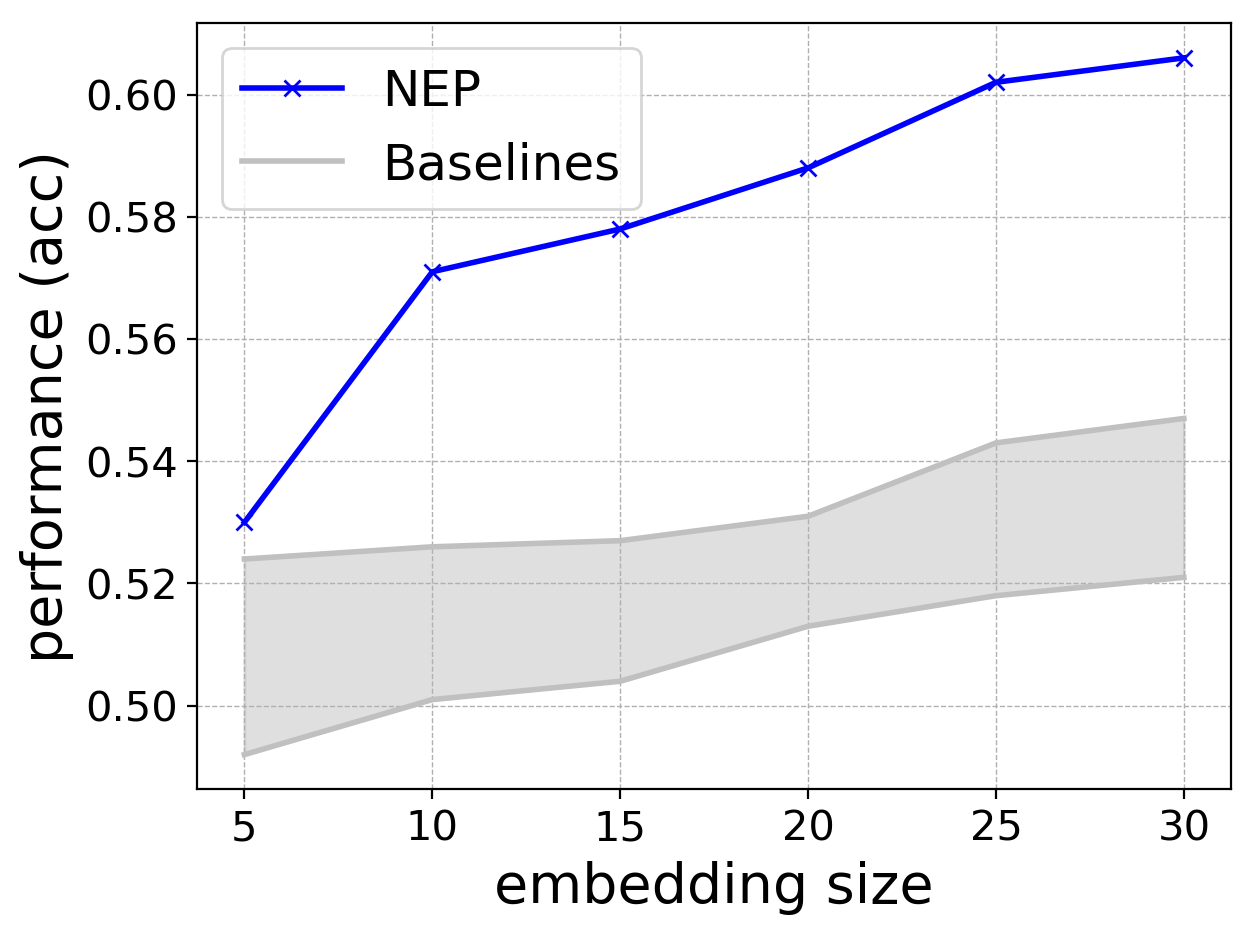}}
\subfigure{
\includegraphics[width=0.29\textwidth]{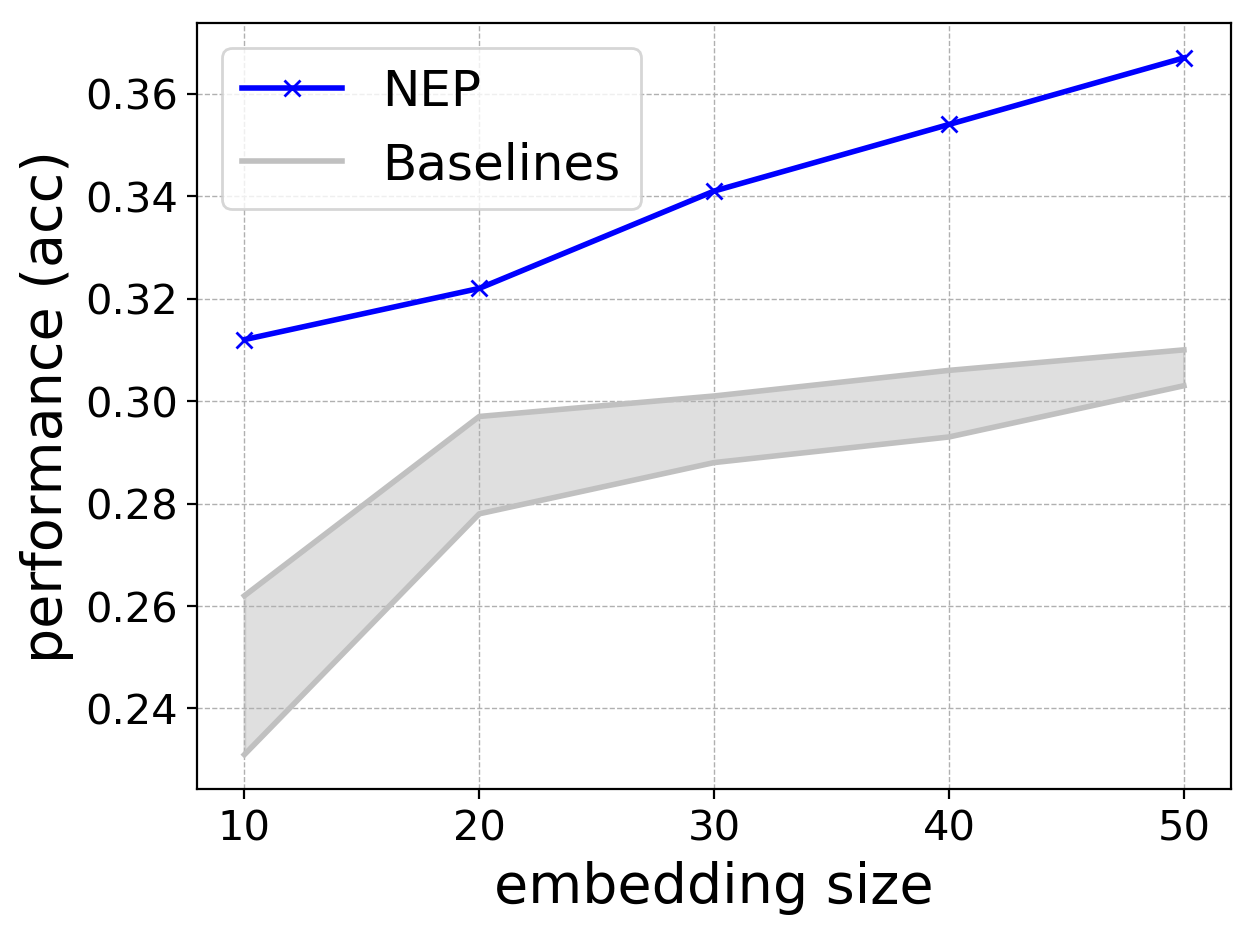}}
\vspace{-5pt}
\caption{\textbf{Robustness of NEP regarding embedding sizes (from left to right: DBLP, YAGO, GitHub).}}
\label{fig:emb}
\end{figure*}

\begin{figure*}[h]
\centering
\subfigure{
\includegraphics[width=0.29\textwidth]{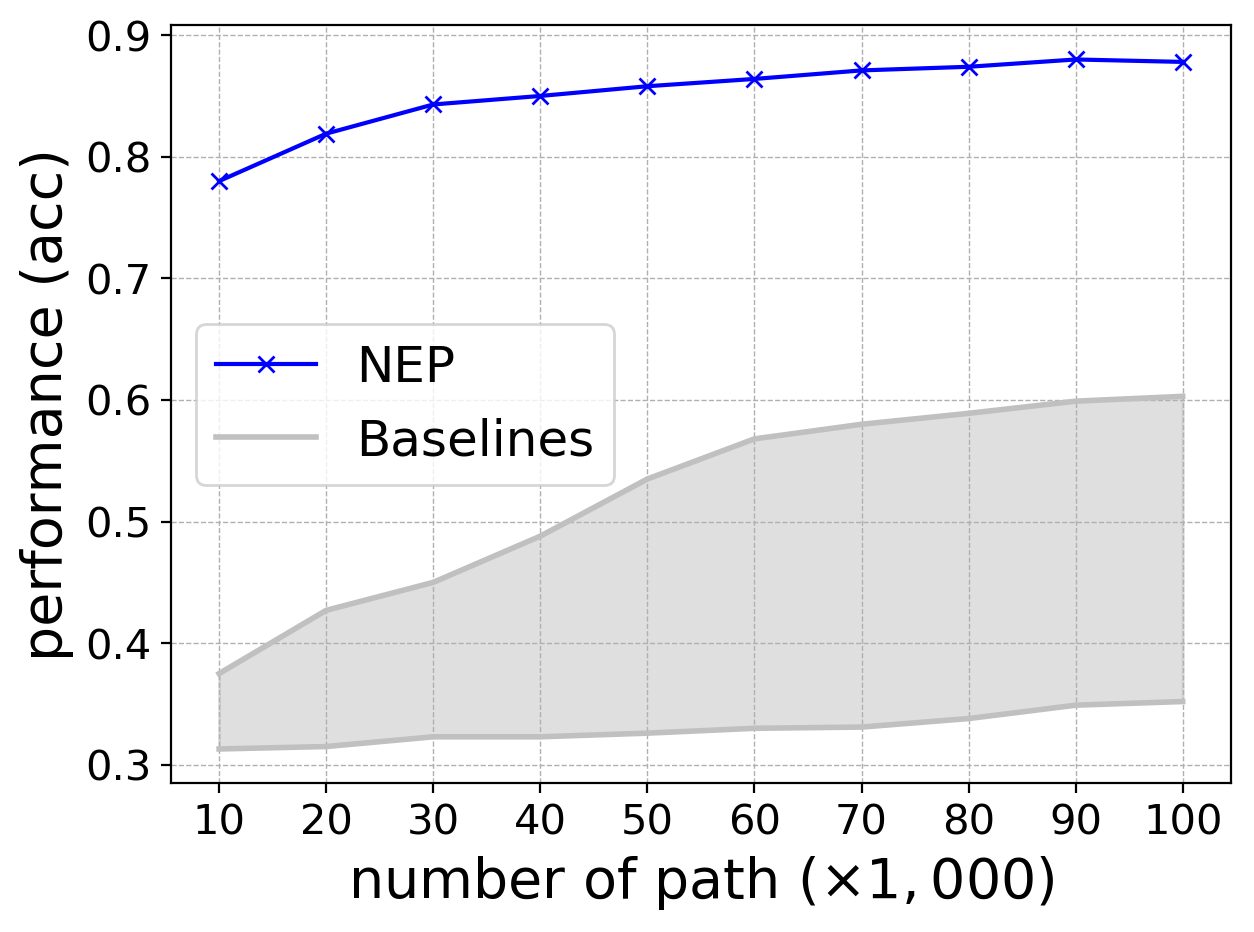}}
\subfigure{
\includegraphics[width=0.29\textwidth]{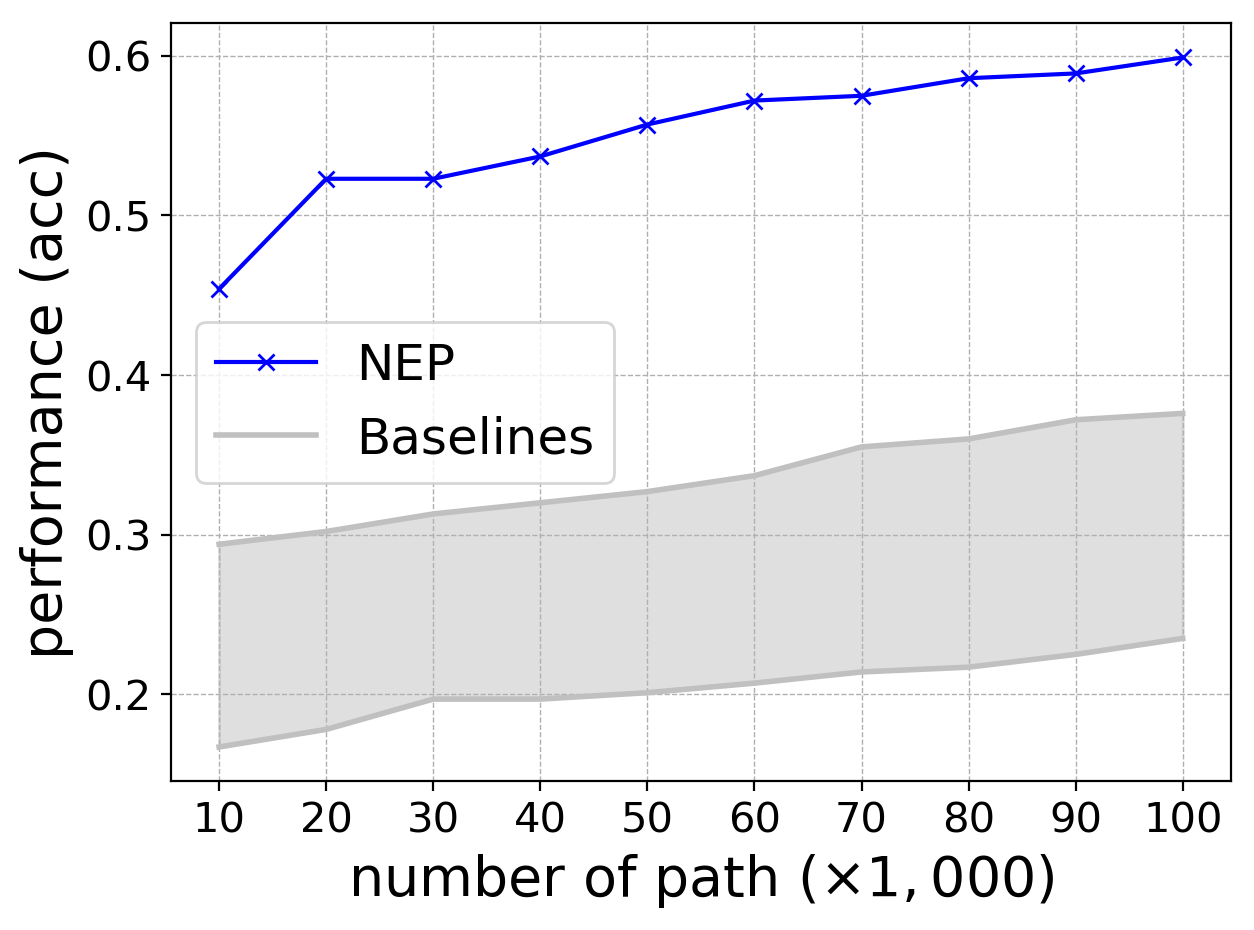}}
\subfigure{
\includegraphics[width=0.29\textwidth]{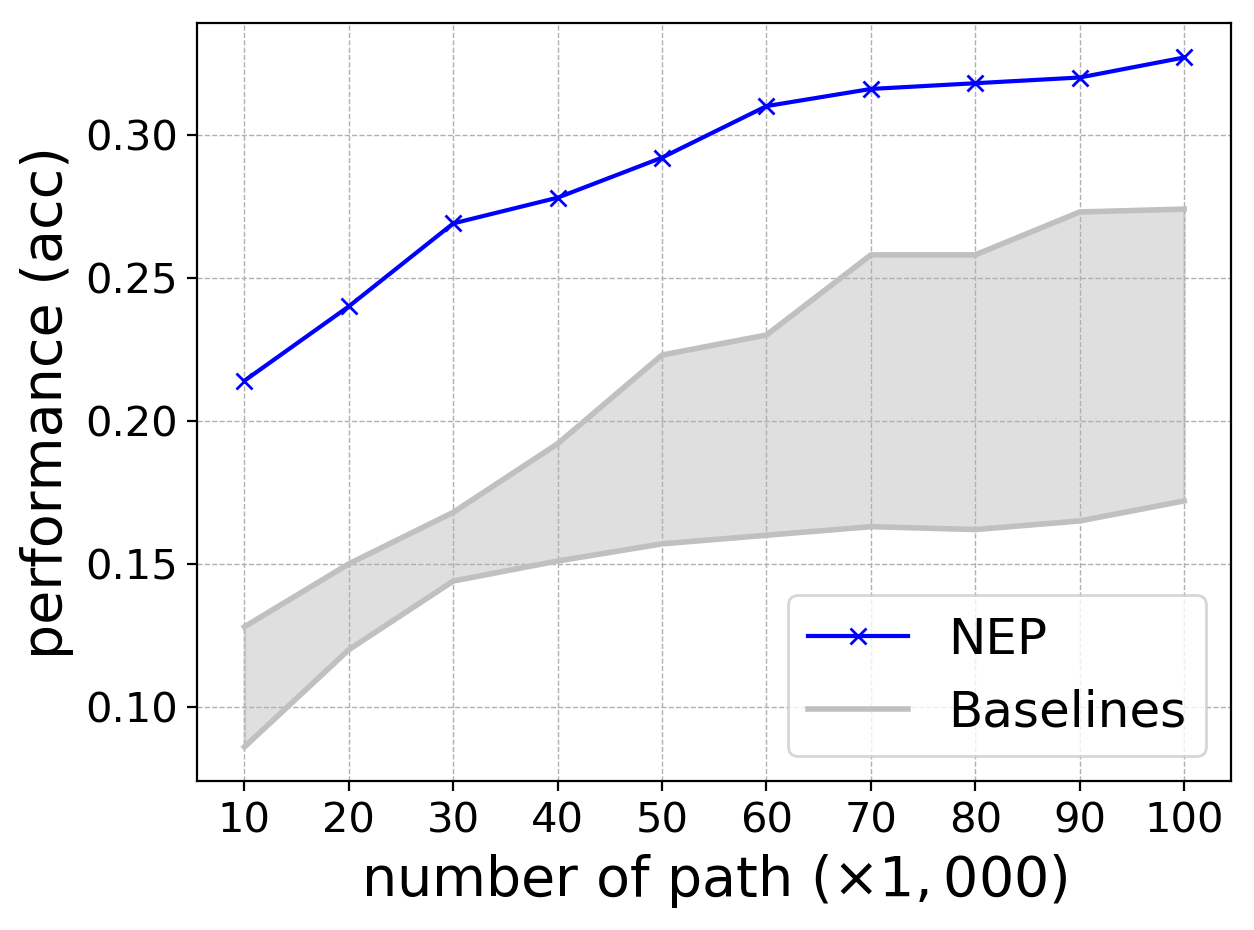}}
\vspace{-5pt}
\caption{\textbf{Robustness of NEP regarding numbers of sampled paths (from left to right: DBLP, YAGO, GitHub).}}
\vspace{5pt}
\label{fig:path}
\end{figure*}

Then we look at the embedding size $K$, by comparing NEP with GHE, Metapath2vec, ESim and Hin2vec, which also compute object embeddings. As shown in Figure \ref{fig:emb}, too small embedding sizes often lead to poor performance. As the embedding size grows, NEP quickly reaches the peak performance. It also maintains the best performance without overfitting the data as the embedding size further grows. 

Finally we look at the number of total sampled paths $\Omega$, by comparing NEP with Metapath2vec, ESim and Hin2vec, which are also trained with path sampling. As shown in Figure \ref{fig:path}, the improvement of NEP over compared baselines is more significant given fewer sampled paths, indicating the power of NEP to rapidly capture useful information in the networks. 

\begin{table}[h]
\small
\centering
 \resizebox{0.48\textwidth}{!}{\begin{tabular}{|c|cccccc|}
   \hline
{\bf Dataset}&{\bf 2}&{\bf 3}&{\bf 4}&{\bf 5}&{\bf 6}&{\bf 7}\\
  \hline
{\bf DBLP}& 0.750 & 0.872 & 0.869 & \underline{0.880}& 0.873 & 0.875\\
\hline
{\bf YAGO}& 0.631 & 0.631 & 0.633 & 0.629& \underline{0.634} & 0.631 \\
\hline
{\bf GitHub}& 0.381 & 0.374 & 0.378 & 0.386 & \underline{0.392}& 0.379\\
\hline
{\bf sub-DBLP}& 0.763 & 0.875 & 0.880 & 0.886& 0.881 & \underline{0.888}\\
\hline
{\bf sub-YAGO}& 0.646 & 0.641 & 0.641 & \underline{0.651}& 0.637 & 0.648\\
\hline
{\bf sub-GitHub}& \underline{0.425} & 0.412 & 0.414 & 0.412& 0.405 & 0.407\\
\hline
 \end{tabular}}
\caption{ \label{tab:length}\textbf{Robustness of NEP regarding path lengths.}}
\end{table}

\section{Conclusions}
\label{sec:con}
In this work, we develop NEP (\textit{Neural Embedding Propagation}), for semi-supervised learning over heterogeneous networks. NEP is a powerful yet efficient neural framework that coherently combines an object encoder and a modular network to model the complex interactions among multi-typed multi-relational objects in heterogeneous networks. Unlike existing heterogeneous network models, NEP does not assume a given set of useful meta-paths, but rather dynamically composes and estimates the different importance and functions of arbitrary meta-paths regarding embedding propagation on-the-fly. At the same time, the model is easy to learn, since the parameters modeling each type of links are shared across all underlying meta-paths.
For future works, it is straightforward to extend NEP to various object attributes and enable fully unsupervised training for by recovering different types of links.


\section*{Acknowledgements}\label{sec:ack}
Research was sponsored in part by U.S. Army Research Lab. under Cooperative Agreement No. W911NF-09-2-0053 (NSCTA), DARPA under Agreement No. W911NF-17-C-0099, National Science Foundation IIS 16-18481, IIS 17-04532, and IIS-17-41317, DTRA HDTRA11810026, and grant 1U54GM114838 awarded by NIGMS through funds provided by the trans-NIH Big Data to Knowledge (BD2K) initiative (www.bd2k.nih.gov). 
\bibliographystyle{IEEEtran}
\small
\bibliography{carlyang} 
\end{document}